\documentclass{article}

\usepackage{microtype}
\usepackage{graphicx}
\usepackage{booktabs} 

\usepackage{hyperref}

\usepackage{amsmath,amsfonts,bm}

\newcommand{\Rnum}[1]{\lowercase\expandafter{\romannumeral #1\relax}}

\def\Figref#1{Figure~\ref{#1}}

\def\Tabref#1{Table~\ref{#1}}

\def\Secref#1{Section~\ref{#1}}

\def\eqref#1{equation~\ref{#1}}

\def\0{\bm{0}} 
\def\1{\bm{1}}

\def\valpha{{\bm{\alpha}}} 
\def\vtheta{{\bm{\theta}}}

\def\vp{{\bm{p}}}
\def\vq{{\bm{q}}}

\def\vx{{\bm{x}}}
\def\vy{{\bm{y}}}

\DeclareMathAlphabet{\mathsfit}{\encodingdefault}{\sfdefault}{m}{sl}
\SetMathAlphabet{\mathsfit}{bold}{\encodingdefault}{\sfdefault}{bx}{n}

\def\sB{{\mathbb{B}}}

\def\sR{{\mathbb{R}}}

\DeclareMathOperator*{\argmax}{arg\,max}

\usepackage{amsmath}
\usepackage{amssymb}
\usepackage[utf8]{inputenc}
\usepackage{subcaption} 
\usepackage{hhline}
\usepackage{graphics}
\usepackage{amsthm}
\usepackage{multirow}
\usepackage{color,soul}
\usepackage{lipsum}

\newcommand\numberthis{\addtocounter{equation}{1}\tag{\theequation}}
\renewcommand{\eqref}[1]{(\ref{#1})} 
\newtheorem{theorem}{Theorem}

\newtheorem{lemma}{Proposition}
\newtheorem{asu}{Assumption}

\newcommand{\RNum}[1]{\uppercase\expandafter{\romannumeral #1\relax}}

\newtheorem*{varthm+inner}{\varthmname}
\newcommand{\varthmname}{}
\newenvironment{varthm}[2][Theorem]
 {\renewcommand{\varthmname}{#1 \ref{#2} (restated)}\begin{varthm+inner}}
 {\end{varthm+inner}}
 
\newtheorem*{varasu+inner}{\varasmname}
\newcommand{\varasmname}{}

\newtheorem*{varlem+inner}{\varlemname}
\newcommand{\varlemname}{}
\newenvironment{varlem}[1]
 {\renewcommand{\varlemname}{Proposition \ref{#1} (restated)}\begin{varlem+inner}}
 {\end{varlem+inner}}

\newlength\mylen

\usepackage{xcolor,colortbl}
\definecolor{Gray}{gray}{0.85}
\newcolumntype{g}{>{\columncolor{Gray}}c}  

\usepackage[accepted]{icml2021}

\icmltitlerunning{Bridged Adversarial Training}

\begin{document}

\twocolumn[
\icmltitle{Bridged Adversarial Training}



\icmlsetsymbol{equal}{*}

\begin{icmlauthorlist}
\icmlauthor{Hoki Kim}{equal,snu}
\icmlauthor{Woojin Lee}{equal,snu}
\icmlauthor{Sungyoon Lee}{equal,snu}
\icmlauthor{Jaewook Lee}{snu}
\end{icmlauthorlist}

\icmlaffiliation{snu}{Seoul National University, Seoul, Korea}

\icmlcorrespondingauthor{Jaewook Lee}{jaewook@snu.ac.kr}





\vskip 0.3in
]



\printAffiliationsAndNotice{\icmlEqualContribution} 

\begin{abstract}
Adversarial robustness is considered as a required property of deep neural networks. In this study, we discover that adversarially trained models might have significantly different characteristics in terms of margin and smoothness, even they show similar robustness. Inspired by the observation, we investigate the effect of different regularizers and discover the negative effect of the smoothness regularizer on maximizing the margin. Based on the analyses, we propose a new method called bridged adversarial training that mitigates the negative effect by bridging the gap between clean and adversarial examples. We provide theoretical and empirical evidence that the proposed method provides stable and better robustness, especially for large perturbations.
\end{abstract}

\section{Introduction}
\label{sec:intro}

Deep neural networks are vulnerable to adversarial examples, which are intentionally perturbed to cause misclassification \cite{szegedy2013intriguing}. 
Since deep neural networks can be applied to various fields, defense techniques against adversarial attacks are now considered an important research area. 
To improve the robustness of neural networks against adversarial attacks, many defense methods have been proposed \cite{goodfellow2014explaining, madry2017towards, tramer2017ensemble, zhang2019theoretically}. Among these, adversarial training (AT) \cite{madry2017towards} and TRADES \cite{zhang2019theoretically} are considered powerful base methods to achieve high adversarial robustness \cite{gowal2020uncovering, wu2020adversarial}. In this paper, while AT and TRADES have similar robustness, we discover that they have totally different \textit{margin} and \textit{smoothness}. 

\begin{figure}
    \centering
    \includegraphics[width=0.9\linewidth]{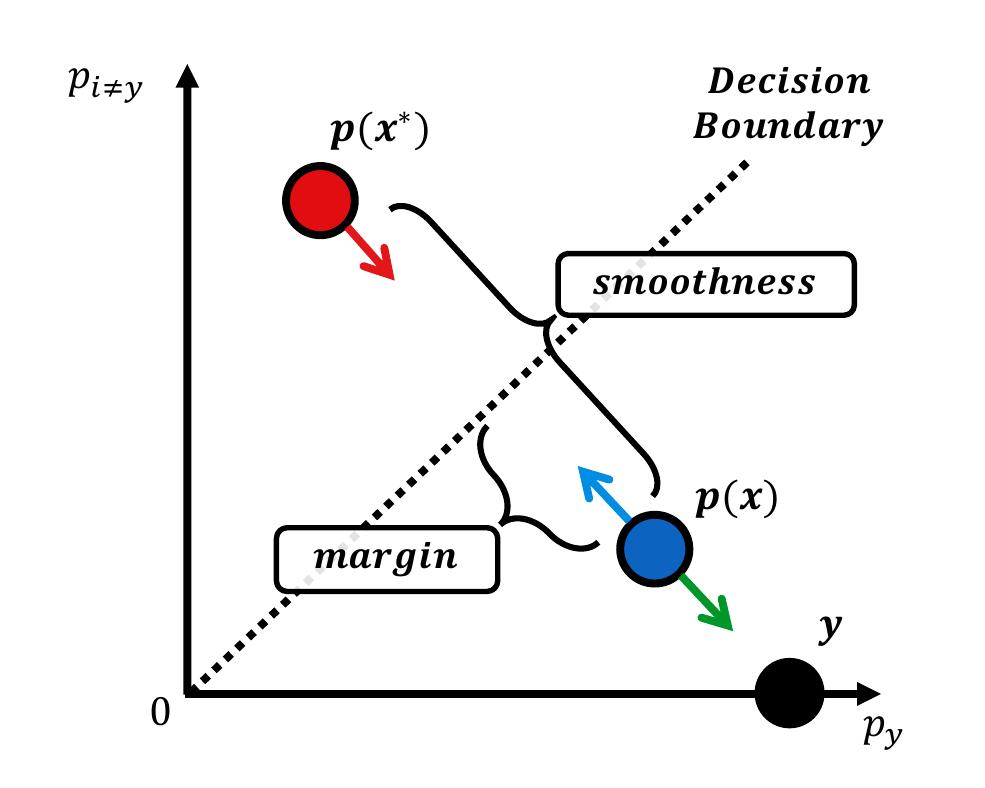}
    \caption{Simple illustration of margin and smoothness in adversarial training. The distance to the decision boundary from an example corresponds to margin. In contrast, the distance between outputs of a clean example $\vx$ and an adversarial example $\vx^*$ corresponds to smoothness. A smaller distance indicates a better smoothness.}
    \label{fig:intro}
\end{figure}

Margin, in general, corresponds to the distance from an example to the decision boundary. For example, given a clean example $\vx$ and its probability output $\vp(\vx)$, the adversarial margin can be defined as the difference between the probability with respect to the true label $y$ and the other most probable class, $\vp(\vx)_y - \max_i \vp(\vx)_{i\neq y}$ \cite{carlini2017towards} as shown in \Figref{fig:intro}. Larger distance indicates better margin. AT tries to maximize the margin of an adversarial example $\vx^*$, which corresponds to the red arrow in \Figref{fig:intro}.

Smoothness corresponds to the insensitiveness of the output to the input perturbation. The $\mathcal{L}_2$ distance $||\vp(\vx) - \vp(\vx^*)||_2$ \cite{kannan2018adversarial} or the Kullback-Leibler divergence $\text{KL}(\vp(\vx)||\vp(\vx^*))$ \cite{zhang2019theoretically} can be easily used to estimate smoothness. TRADES tries to maximize the margin of an clean example $\vx$ (green arrow), while minimize the smoothness between $\vp(\vx)$ and $\vp(\vx^*)$ (red and blue arrows).

Inspired by the observation, we investigate the characteristics of the regularizers of AT and TRADES, and find that there exists the negative effect of the smoothness regularizer on maximizing the margin. From the analyses, we propose a novel method to mitigate the negative effect and provide stable performance by bridging the gap between clean and adversarial examples.

\section{Related Work and Background}
\subsection{Notations}
We consider a $c$-class classification task with a neural network $f_{\vtheta}:\mathcal{X} \rightarrow \mathbb{R}^c$.
The network $f_\vtheta$ classifies a sample $\vx\in\mathcal{X}$ as $\argmax_{i\in\mathcal{Y}} [f_\vtheta(\vx)]_i$, where $\mathcal{Y}=\{0, \cdots, c-1\}$.
We denote the true label with respect to $\vx$ by $y$ and the corresponding one-hot representation by $\vy\in\{0,1\}^c$. That is, $y_i=\1\{i=y\}, \forall i\in\mathcal{Y}$, with an indicator function $\boldsymbol{1}\{C\}$ which outputs 1 if the condition $C$ is true and 0 otherwise.
Then the probability function $\vp_{\vtheta} = \text{softmax}\circ f_{\vtheta}:\mathcal{X}\rightarrow [0,1]^c$ outputs a $c$-dimensional probability vector whose elements sum to 1.

Given two probability vectors $\vp,\vq$ in the $c$-dimensional probability simplex, we define the following values:
$H_\vp(\vq) = -\vp^T \log \vq $ and $\text{KL}(\vp||\vq) = \vp^T \log \frac{\vp}{\vq} $.
These are called the cross-entropy and Kullback-Leibler (KL) divergence between $\vp$ and $\vq$, respectively.
In addition, we denote the entropy of $\vp$ as $H(\vp) = H_\vp(\vp) =  -\vp^T \log \vp $.
Note that for a one-hot vector $\vy\in\{0,1\}^c$, $\text{KL}(\vy||\vq) = \sum_{i\in \mathcal{Y}} y_i \log \frac{y_i}{q_i} =  -\vy^T \log \vq $ is equivalent to the well-known cross-entropy, $H_\vy(\vq)$.

\subsection{Adversarial Robustness}

Since \citet{szegedy2013intriguing} identified the existence of adversarial examples, most defenses are broken by adaptive attacks \cite{athalye2018obfuscated, tramer2020adaptive} and the state-of-art performance is still observed from variants of adversarial training \cite{madry2017towards} and TRADES \cite{zhang2019theoretically} utilizing the training tricks \cite{pang2020bag, gowal2020uncovering}, weight averaging \cite{wu2020adversarial}, and using more data \cite{carmon2019unlabeled, rebuffi2021fixing}.

\textbf{Adversarial Training (AT)} \cite{madry2017towards}
is one of the most effective defense methods. Given a perturbation set $\sB(\vx,\epsilon)$, which denotes a ball around an example $\vx$ with a maximum perturbation $\epsilon$, it encourages the worst-case probability output over the perturbation set $\sB(\vx,\epsilon)$ to directly match the label $\vy$ by minimizing the following loss:
\begin{equation} \label{eq:AT}
\begin{split}
    \ell_{AT}(\vx,\vy;\vtheta)=&\max_{\vx' \in \sB(\vx,\epsilon)}H_\vy(\vp_\vtheta(\vx'))\\
    =&\max_{\vx' \in \sB(\vx,\epsilon)}\text{KL}(\vy||\vp_\vtheta(\vx')).
\end{split}
\end{equation}
 
\textbf{TRADES} \cite{zhang2019theoretically} was proposed based on the analysis of the trade-off between adversarial robustness and standard accuracy. TRADES minimizes the following loss:
\begin{equation} \label{eq:TRADES}
\begin{split}
    &\ell_{TRADES}(\vx,\vy;\vtheta)\\
    =&\text{KL}(\vy||\vp_\vtheta(\vx))+\beta\max_{\vx' \in \sB(\vx,\epsilon)} \text{KL}(\vp_\vtheta(\vx)||\vp_\vtheta(\vx')).
\end{split}
\end{equation}
where $\beta$ is the regularization hyper-parameter. Here, the first term aims to maximize the margin of clean examples, while the second term encourages the model to be smooth.

To solve this highly non-concave optimization in \eqref{eq:AT} and \eqref{eq:TRADES}, an iterative projected gradient descent (PGD) with $n$ steps is widely used:
\begin{equation}
    \vx^{t+1} = \Pi_{\sB(\vx,\epsilon)} \big(\vx^t + \alpha\cdot\text{sign} (\nabla_\vx \ell_{inner}(\vx, \vy))\big)
\end{equation}
where $\Pi_{\sB(\vx,\epsilon)}$ refers the projection to the $\sB(\vx,\epsilon)$ and $\alpha$ is a step size for each step. Here, $\vx^0$ is the original example and $\vx^n$ is used an adversarial example $\vx^*$. We denote this as PGD$^n$. For example, AT aims to minimize the loss in \eqref{eq:AT} so that $\text{KL}(\vy||\vp_\vtheta(\vx^t))$ is used as $\ell_{inner}(\vx, \vy)$. 

\subsection{Margin and Smoothness}

To achieve a higher accuracy, margin and smoothness have been considered as important characteristics of deep neural networks \cite{elsayed2018large, sokolic2017robust, anil2019sorting, fazlyab2019efficient}. Following prior works, the concept of margin and smoothness also has been adopted in the adversarial training framework. 

In the case of margin, max-margin adversarial training (MMA) \cite{ding2019mma} trains adversarial examples for the correctly classified examples, and clean examples for the misclassified examples to maximize the input space margin, which is the distance to the decision boundary in the input space. \citet{wang2019improving} outperformed MMA by emphasizing the regularization of the misclassified examples. Note that \citet{wang2019improving} used the output space margin that is the distance to the decision boundary in the output space, which we use in this paper. \citet{sanyal2020benign} discovered that adversarial training models have a larger margin than naturally trained models, and connected it to the complexity of decision boundaries. \citet{yang2020boundary} focused on the boundary thickness, an extended concept of the margin, and connected it to adversarial robustness.

In the case of smoothness, although there are several methods such as the Parseval network \cite{cisse2017parseval}, input gradient regularization \cite{ross2017improving}, and adversarial logit pairing \cite{kannan2018adversarial}, TRADES outperforms any other methods. \citet{hein2017formal} connected the instance-based local Lipschitz to adversarial robustness. \citet{yang2020closer} also concluded that local Lipschitzness is correlated with adversarial robustness.

However, while the correlation between the margin and smoothness in standard training has been discussed \cite{von2004distance, xu2009robustness}, none of the works analyzed both margin and smoothness together in the adversarial training framework. Note that provable defensive methods have discussed the trade-off between the margin and smoothness \cite{salman2019provably, chen2020efficient}, however, they are in a different direction from the adversarial training frameworks. Thus, we analyze the margin and smoothness of different adversarial training frameworks, and connect it to their regularization terms. 

\section{Understanding Margin and Smoothness in Adversarial Training} \label{sec:motivation}
We will start by introducing the difference in margin and smoothness between AT and TRADES. Then, we explore the cause of the difference by analyzing their regularizers.

\subsection{Similar robustness, but different margin and smoothness}
To illustrate the difference between AT and TRADES in terms of margin and smoothness, we first define a measure for margin and smoothness. To estimate margin, we use $M(\cdot)$ following \cite{carlini2017towards}:
\begin{equation}\label{eq:margin}
    M(\vx): =\vp_\vtheta(\vx)_y-\max_{i\neq y}\vp_\vtheta(\vx)_i
\end{equation}
Thus, $M(\vx)>0$ indicates that the model correctly predicts the label of $\vx$. On the contrary, the model outputs a wrong prediction when $M(\vx)<0$. To estimate smoothness, we use $\text{KL}(\vp_\vtheta(\vx)||\vp_\vtheta(\vx^*))$ in \cite{zhang2019theoretically}, where $\vx^*$ is an adversarial example of a clean example $\vx$. We note that sliced Wasserstein distance and Jensen-
Shannon divergence also can be used to measure smoothness, and we observed that the overall results are similar to the ones with the KL divergence.

\begin{figure}
    \centering
    \begin{subfigure}[ht]{\columnwidth}
      \centering
        \includegraphics[width=0.95\linewidth]{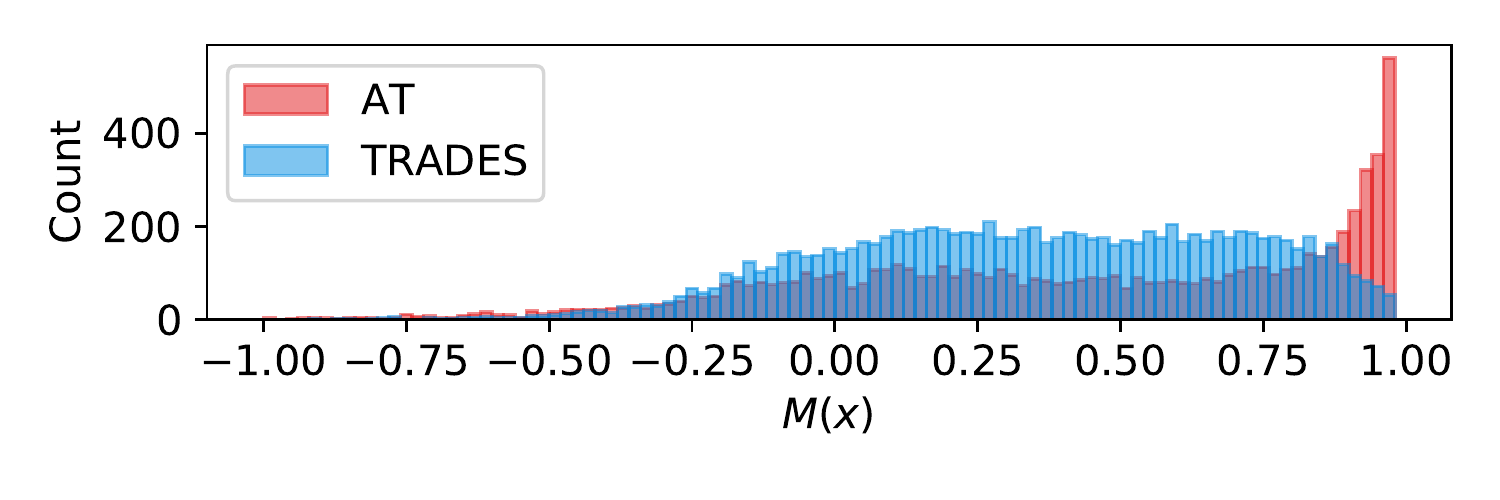}
      \caption{Margin}
        \label{fig:margin}  
    \end{subfigure}        
    
    \begin{subfigure}[ht]{\columnwidth}
      \centering
        \includegraphics[width=0.95\linewidth]{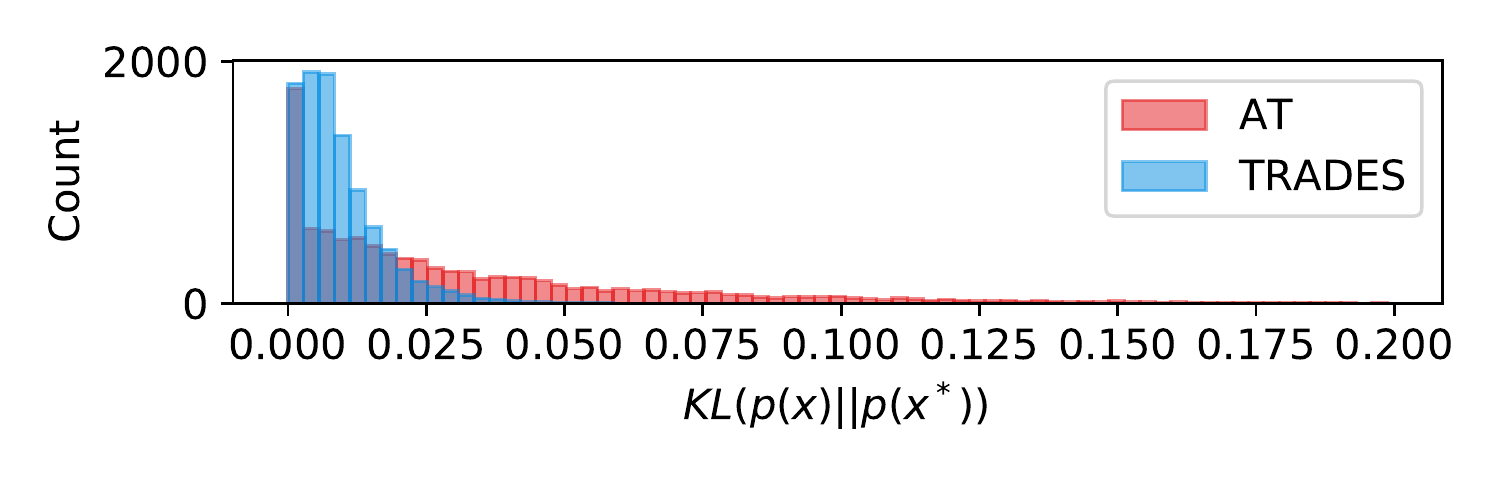} 
    \caption{Smoothness}
        \label{fig:smoothness}  
    \end{subfigure}     
    \caption{Margin and smoothness of AT and TRADES. (a) $M(\vx)$ for estimating margin (higher is better). (b) $\text{KL}(\vp_\theta(\vx)||\vp_\theta(\vx^*))$ for estimating smoothness (lower is better). Each plot used 10,000 test examples. Although they show similar robustness, the characteristics are entirely different.
    }
    \label{fig:marginsmoothness}
\end{figure}

\Figref{fig:marginsmoothness} illustrates the difference between AT and TRADES in terms of margin and smoothness. We generated adversarial examples with PGD$^{50}$ for models trained on CIFAR10. Detailed settings are presented in \Secref{sec:Experiments}. Although AT and TRADES have similar robustness, they show totally different characteristics. AT shows a larger margin that is distributed close to 1, whereas it has a poor smoothness than TRADES.
On the contrary, TRADES shows a smaller margin with only a few examples around 1, whereas it has a better smoothness than AT.

\subsection{Effect of regularizers for margin and smoothness}

\begin{figure}
    \centering
    \includegraphics[width=\linewidth]{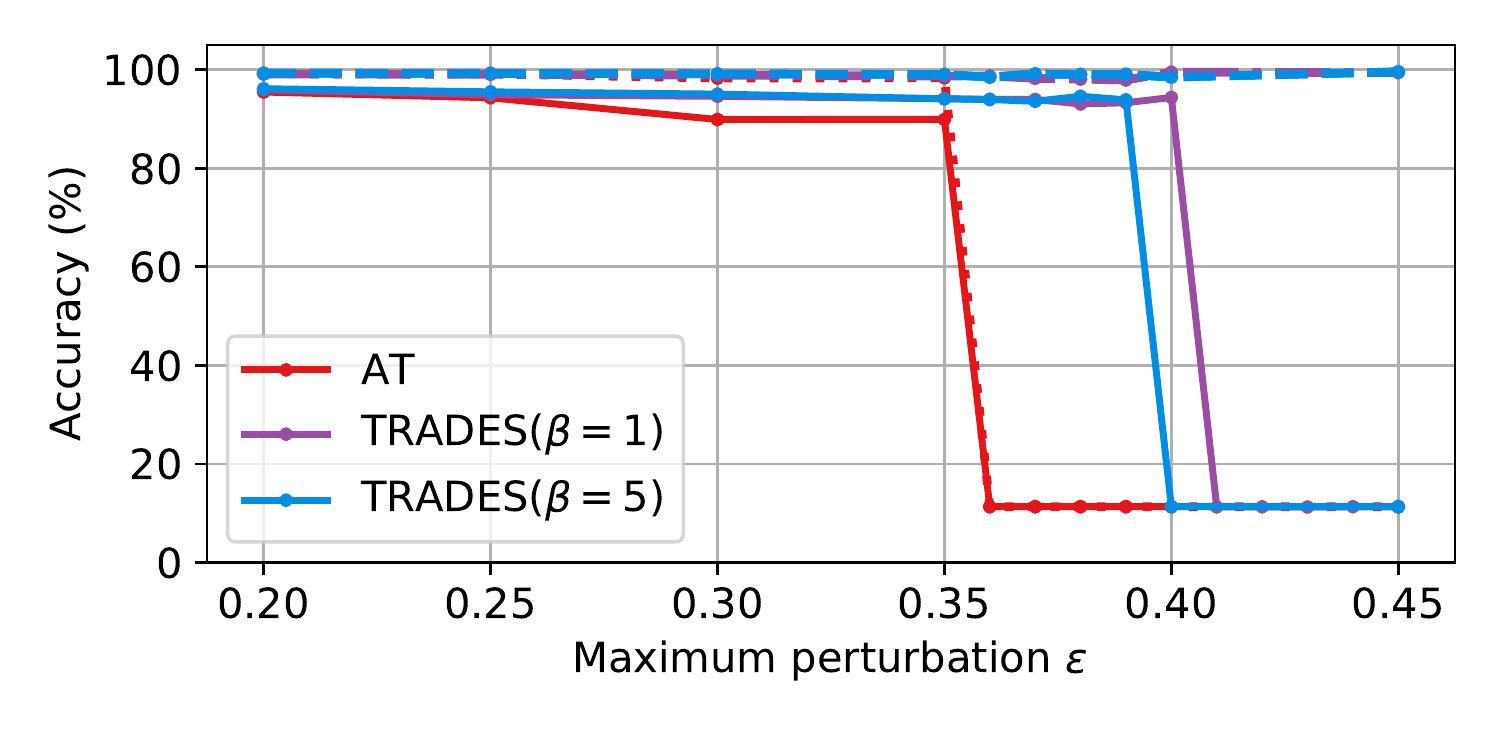}
    \caption{Accuracy of each training method with a wide range of the maximum perturbation $\epsilon$. Each model is trained on the given maximum perturbation $\epsilon$ and evaluated with the same $\epsilon$. The dotted and solid lines indicates the clean accuracy and robust accuracy, respectively. 
    \label{fig:mnist_eps}
    }
\end{figure}

To analyze the cause of these different characteristics, we take a closer look at the regularization terms of AT and TRADES. First, AT directly increases the margin of adversarial examples $\vx^*$ as in \eqref{eq:AT}. However, as there is no loss term for controlling the distance between clean and adversarial examples, AT is difficult to attain smoothness, which is observed in \Figref{fig:marginsmoothness}. Moreover, it is recently discovered that the regularization term for maximizing the margin of adversarial examples has some drawbacks in convergence \cite{shaeiri2020towards, liu2020loss, dong2021exploring, sitawarin2020improving, shaeiri2020towards}. Indeed, following \cite{sitawarin2020improving, shaeiri2020towards}, when we evaluate the robustness against a wide range of the maximum perturbation $\epsilon$ on MNIST, AT fails to achieve sufficient standard and robust accuracy for $\epsilon > 0.35$ as shown in \Figref{fig:mnist_eps}.

In contrast, TRADES adopts the regularization term $\text{KL}(\vp_\vtheta(\vx)||\vp_\vtheta(\vx'))$ for smoothness in \eqref{eq:TRADES}. By doing so, TRADES gains a better smoothness as shown in \Figref{fig:marginsmoothness}. However, TRADES fails to achieve a high margin even though TRADES has a regularization term for maximizing the margin. Another interesting point regards a poor adversarial robustness and high clean accuracy for $\epsilon \geq 0.4$ on MNIST in \Figref{fig:mnist_eps} even with a smaller weight $\beta=1$. In summary, TRADES has advantages compared to AT in that it shows more stable performance with high clean accuracy, but still has trouble optimizing \eqref{eq:TRADES} which maximizes the margin and minimizes the KL divergence simultaneously. 

\subsection{Negative effect of smoothness regularizer on maximizing the margin}
The degraded margin of TRADES and its failure cases for a larger perturbations lead us to postulate the hypothesis that there is a conflict between $\text{KL}(\vy||\vp_\vtheta(\vx))$ and $\text{KL}(\vp_\vtheta(\vx)||\vp_\vtheta(\vx^*))$. Now, we mathematically prove that the regularizer for smoothness $\text{KL}(\vp_\vtheta(\vx)||\vp_\vtheta(\vx^*))$ in \eqref{eq:TRADES} has a negative effect on training a large margin. This is formalized in the following proposition.

\begin{figure}
    \centering
    \includegraphics[width=0.9\linewidth]{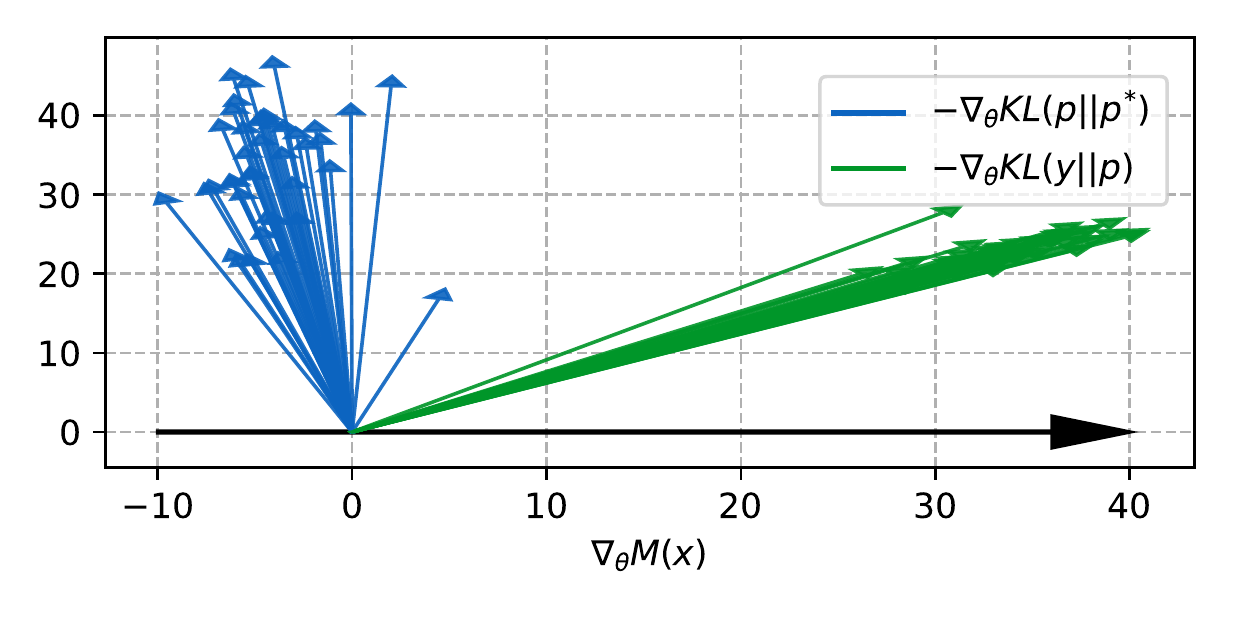}
    \caption{Gradients of each loss term of TRADES in \eqref{eq:TRADES}. The direction of each vector is denoted as the angle to the gradient of increasing the margin at the end of each epoch.
    Except for first two epochs, each terms' gradients are opposite to each other for the 3rd epoch to the 40th epoch.
    \label{fig:grad}
    }
\end{figure}

\begin{figure}
    \centering
    \begin{subfigure}[ht]{\columnwidth}
      \centering
      \includegraphics[width=0.9\linewidth]{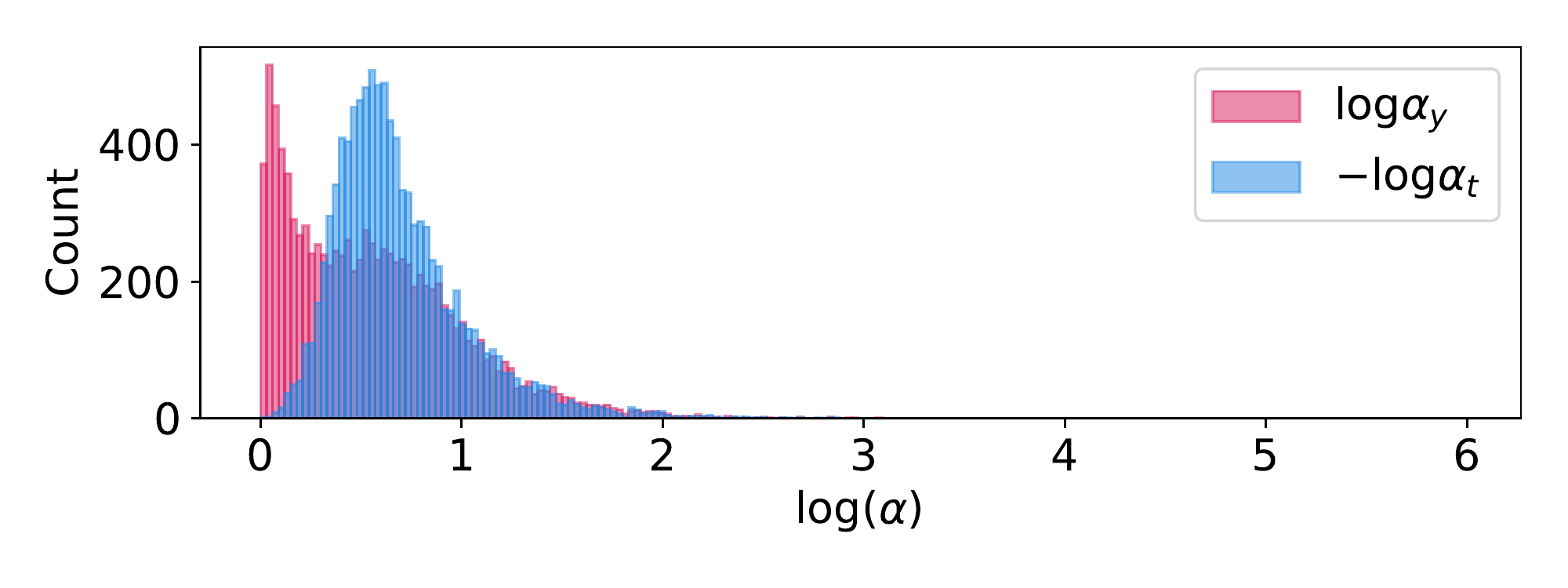}
      \caption{$\log \alpha_i$ for $\epsilon=8/255$}
      \label{fig:alpha_eps8}
    \end{subfigure}        
    \begin{subfigure}[ht]{\columnwidth}
      \centering
      \includegraphics[width=0.9\linewidth]{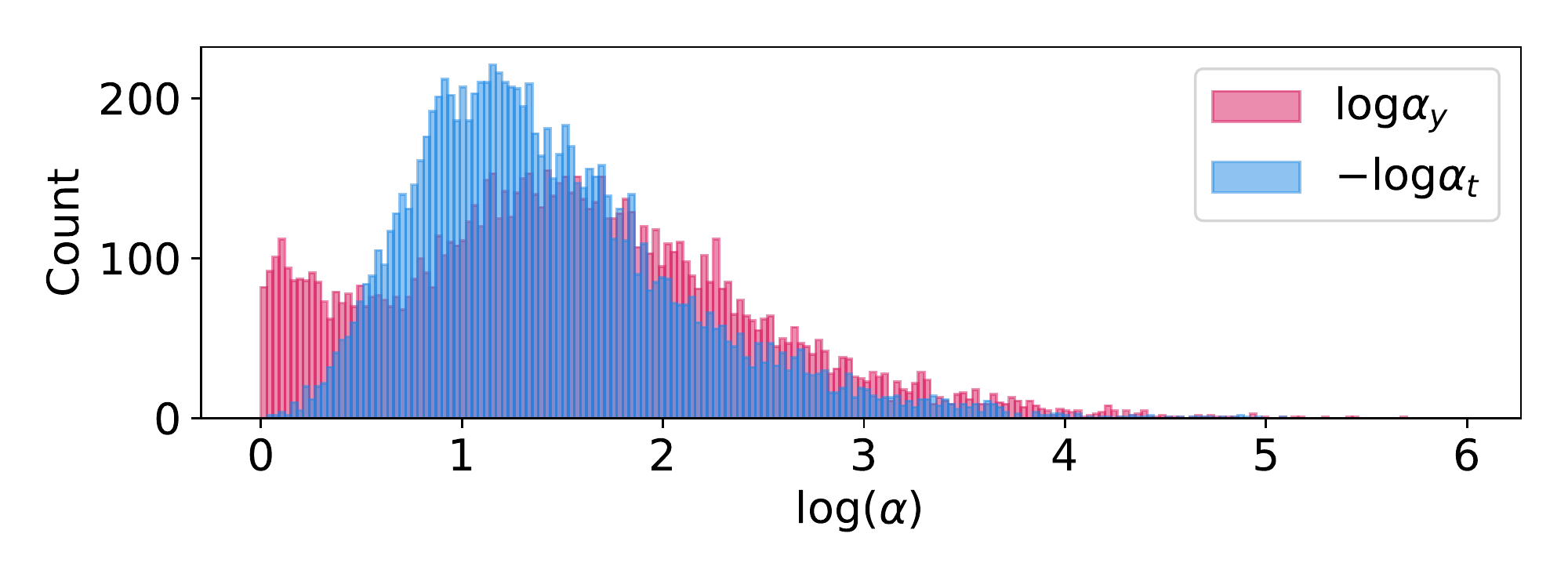}
      \caption{$\log \alpha_i$ for $\epsilon=16/255$}
      \label{fig:alpha_eps16}
    \end{subfigure}        
    \caption{Distribution of 
    $\log \alpha_y=\log (p_y/p^*_y)$ for the true class $y$ (red) and $-\log \alpha_t=-\log (p_t/p^*_t)$ for the target class $t = \arg\max_{i\neq y} p^*_i$ (blue). Each plot shows the distribution of $\log \alpha_y$ and $\log \alpha_t$ for the given $\vp^*$ under the maximum perturbation $\epsilon$.
    }
    \label{fig:alpha}
\end{figure}

\begin{lemma}\label{lem:gradient}
Let $\vp=\vp_\vtheta(\vx)$, $\vp^*=\vp_\vtheta(\vx^*)$, $\nabla=\nabla_\vtheta$, and $t=\max_{i\neq y}\vp$. If $\log\alpha_y>0$ and $\log\alpha_t <0$ for $\log \alpha_i =\log\frac{p_i}{p^*_i}$, then the gradient descent direction of 
$\text{KL}(\vp||\vp^*)$ is aligned with the gradient direction that minimizes the margin $M(\vx)=p_y-p_t$ by penalizing $p_i$ with the scale of $\log \alpha_i$.
\begin{align*}
    -\nabla \text{KL}(\vp||\vp^*) = - (\nabla p_y)^T \log\alpha_y - (\nabla p_t)^T \log\alpha_t + c
\end{align*}
where $c$ is a linear combination of other gradient directions.
\end{lemma}
The proof is provided in Appendix \ref{ap:proof}. Note that the assumption, $\log\alpha_y>0$ and $\log\alpha_t <0$, is generally acceptable under the characteristic of adversarial attack as shown in \Figref{fig:alpha}. The proposition tells us that the regularization term for smoothness $\text{KL}(\vp_\vtheta(\vx)||\vp_\vtheta(\vx^*))$ hinders the model from maximizing the margin.

To provide an empirical evidence of the negative effect, we visualize the effect of each loss term of TRADES on the margin in \Figref{fig:grad}. The $x$-axis denotes the gradient direction that maximizes the margin, $\nabla M(\vx)$. The blue and green arrows are obtained using the cosine similarity between $\nabla M(\vx)$. While the gradient descent direction of $\text{KL}(\vy||\vp)$ is aligned with $\nabla M(\vx)$, the gradient descent of the other term $\text{KL}(\vp||\vp^*)$ is in the opposite direction. The result confirms that minimizing the regularization term $\text{KL}(\vp||\vp^*)$ has a negative effect on maximizing the margin.

Moreover, as proved in Proposition \ref{lem:gradient}, the negative effect is proportional to the maximum perturbation $\epsilon$. If we set a larger maximum perturbation $\epsilon=16/255$, the value of $\log\alpha_i$ tends to have a larger deviation from $0$ than that of $\epsilon=8/255$. This can be guaranteed because, for $\epsilon_1 < \epsilon_2$, $\sB(\vx,\epsilon_1) \in \sB(\vx,\epsilon_2)$ so that $\max_{\sB(\vx,\epsilon_1)}\log(\vp_\vtheta(\vx)/\vp_\vtheta(\vx^*)) \leq \max_{\sB(\vx,\epsilon_2)}\log(\vp_\vtheta(\vx)/\vp_\vtheta(\vx^*))$. Thus, minimizing $\text{KL}(\vp||\vp^*)$ with large $\epsilon$ comes with a prohibitive negative effect on maximizing the margin. This is consistent with the fact that TRADES suffers the convergence problem for a larger perturbation in \Figref{fig:mnist_eps}. Thus, from above observation, we expect to enable the model to converge to a better local minima by mitigating the negative effect.

\section{Bridged Adversarial Training} \label{sec:method}
In this section, to mitigate the negative effect of the smoothness regularization term $\text{KL}(\vp||\vp^*)$ on maximizing the margin, we propose a new adversarial training loss. Then, we extend the proposed method and prove that it provides an upper bound on the robust error.

\subsection{Mitigating the negative effect by bridging}
The key idea to alleviate the negative effect of $- (\nabla p_i)^T \log \alpha_i$ is bridging the gap between $\vp$ and $\vp^*$. Suppose we have an intermediate probability $\tilde{\vp}$. Then, the gradient of $\text{KL}(\vp||\tilde{\vp})+\text{KL}(\tilde{\vp}||\vp^*)$ can be calculated by using Proposition \ref{lem:gradient} as follows:
\begin{align*}\label{eq:alpha1}
    &\nabla (\text{KL}(\vp||\tilde{\vp})+\text{KL}(\tilde{\vp}||\vp^*))
    \\ = &  (\nabla \vp)^T \log \valpha^{(1)}
    \\ &+ (\nabla \tilde{\vp})^T \left(-\valpha^{(1)}+\log \valpha^{(2)} \right) - ( \nabla \vp^*)^T \valpha^{(2)}. \numberthis
\end{align*}
where $\alpha_i^{(1)}=\frac{p_i}{\tilde{p_i}}$, $\alpha_i^{(2)}=\frac{\tilde{p_i}}{p^*_i}$, and $\valpha^{(1)},\valpha^{(2)}\in \sR^c$ are the vector whose $i$th elements are $\alpha_i^{(1)}$ and $\alpha_i^{(2)}$, respectively. By controlling $\vert\valpha^{(1)}\vert$ with the intermediate probability $\tilde{\vp}$, we can mitigate the negative effect of the KL divergence loss term. In other words, minimizing the new loss $\text{KL}(\vp||\tilde{\vp})+\text{KL}(\tilde{\vp}||\vp^*)$ achieves the smoothness between $\vp$ and $\vp^*$ with the reduced negative effect on maximizing the margin by using $\tilde{\vp}$ as a bridge. Thus, we name a new adversarial training method, which minimizes $\text{KL}(\vp||\tilde{\vp})+\text{KL}(\tilde{\vp}||\vp^*)$ instead of $\text{KL}(\vp||\vp^*)$, \textit{bridged adversarial training (BAT)}.

\begin{algorithm}[tb]
   \caption{Generalized Bridged Adversarial Training}
   \label{alg:1}
\begin{algorithmic}
   \STATE {\bfseries Input:} training data $\mathcal{D}$; a continuous path $\gamma(\cdot)$; a model with parameter $\vtheta$; an adversarial attack $\mathcal{A}_\vtheta:\mathcal{X}\times\mathcal{Y}\rightarrow\mathcal{X}$; the number of bridge $m$.
   \FOR{$(\vx,\vy) \sim \mathcal{D}$}
   \STATE $\vx^* \leftarrow \mathcal{A}_\vtheta(\vx, \vy)$
   \STATE $\gamma(0) \leftarrow \vx$ and $\gamma(1) \leftarrow \vx^*$
   \STATE $\ell \leftarrow \text{KL}(\vy||\vp_\vtheta(\vx))$
   \STATE $ \quad \quad +\sum_{k=0}^{m-1} \text{KL}\big(\vp_\vtheta(\gamma(\frac{k}{m}))|| \vp_\vtheta(\gamma(\frac{k+1}{m}))\big)$
   \STATE $\vtheta\leftarrow\vtheta-\nabla_\vtheta \ell$
   \ENDFOR
\end{algorithmic}
\end{algorithm}

Intuitively, the more intermediate probabilities induce the less negative effect of $\text{KL}(\vp||\tilde{\vp})$. For a given sample $\vx$, let $\gamma:[0,1]\rightarrow\mathcal{X}$ be a continuous path from $\gamma(0)=\boldsymbol{\vx}$ to $\gamma(1)=\boldsymbol{\vx}^*$, where $\boldsymbol{\vx}^*$ is an adversarial example of $\vx$. Now, we minimize the bridged loss $\sum_{k=0}^{m-1}\text{KL}(\vp_\vtheta(\gamma(\frac{k}{m}))||\vp_\vtheta(\gamma(\frac{k+1}{m})))$ instead of $\text{KL}(\vp_\vtheta(\vx)||\vp_\vtheta(\vx^*))$. Here, $m$ is a hyper-parameter for the number of intermediate probabilities. The generalized bridged adversarial training procedure is presented in Algorithm \ref{alg:1}. Unless otherwise specified, we uses $m=2$, a simple linear path $\gamma(t)=(1-t)\vx+t\vx^*$ for generating the intermediate probability, and the cross-entropy loss as the inner maximization objective. 


\subsection{Bound on the robust error}
Following \citet{zhang2019theoretically}, we provide theoretical evidence that the proposed loss serves as an upper bound on the robust error of the model under the binary classification setting. In the binary classification case, a model can be denoted as $f:\mathcal{X} \rightarrow \mathbb{R}$.
Given a sample $\boldsymbol{x} \in \mathcal{X}$ and a label $y \in \{-1, 1\}$, we use $\text{sign}(f(\boldsymbol{x}))$ as a prediction value of $y$.

Formally, given a surrogate loss $\phi$ and $\eta \in [0,1]$, the conditional $\phi$-risk can be denoted as $H(\eta):=\inf_{\alpha\in\mathbb{R}}(\eta \phi (\alpha) + (1-\eta)\phi(-\alpha))$. Similarly, we can define $H^-(\eta):=\inf_{\alpha(2\eta-1)\leq 0 } (\eta \phi (\alpha) + (1-\eta)\phi(-\alpha))$. Now, we assume the surrogate loss $\phi$ is classification-calibrated, so that $H^-(\eta) > H(\eta)$ for any $\eta \neq 1/2$. Then, the $\psi$-transform of a loss function $\phi$, which is the convexified version of $\hat{\psi}(\theta)=H^-(\frac{1+\theta}{2})-H(\frac{1+\theta}{2})$, is continuous convex function on $\theta \in [-1,1]$.

Then, $\mathcal{R}_{rob}(f):=\mathbb{E}_{(\boldsymbol{x},y)} \boldsymbol{1}\{ \exists \boldsymbol{x'}\in\mathbb{B}(\boldsymbol{x},\epsilon) \text{ s.t. } f(\boldsymbol{x'})y \leq 0 \}$ is the robust error. 
Similarly, $\mathcal{R}_{nat}(f):=\mathbb{E}_{(\boldsymbol{x},y)} \boldsymbol{1}\{f(\boldsymbol{x})y\leq0\}$ is the natural classification error.
Then, $\mathcal{R}_{bdy}(f):=\mathbb{E}_{(\boldsymbol{x},y)} \boldsymbol{1}\{f(\boldsymbol{x})y>0, \exists \vx'\in\sB(\vx,\epsilon) \text{ s.t. } f(\vx)f(\vx')\leq0\}$ is the boundary error by \eqref{eq:decom}.
Given a classification-calibrated surrogate loss function $\phi$ and a surrogate loss $\mathcal{R}_{\phi}(f):= \mathbb{E}_{(\boldsymbol{x},y)} \phi(f(\boldsymbol{x})y)$, the following theorem is demonstrated.

\begin{theorem}\label{th:final}
Given a sample $\vx$ and a positive $\beta$, let $\gamma:[0,1]\rightarrow\mathcal{X}$ be a continuous path from $\gamma(0)=\boldsymbol{x}$ to $\gamma(1)=\boldsymbol{x}^*$ where $\boldsymbol{x}^* = \arg\max_{\boldsymbol{x'} \in \mathbb{B}(\boldsymbol{x}, \epsilon)} \boldsymbol{1}\{\beta f(\boldsymbol{x'})f(\boldsymbol{x})<0\}$. Then, we have
\begin{align*}
    \mathcal{R}_{rob}(f) - \mathcal{R}_{nat}^\star & \leq  \psi^{-1}(\mathcal{R}_{\phi}(f)-\mathcal{R}^\star_\phi)
    \\ &+ \mathbb{E}_{(\boldsymbol{x},y)}\sum_{k=0}^{m-1}\phi(\beta f(\gamma(\frac{k}{m}))f(\gamma(\frac{k+1}{m}))) 
\end{align*}
where $\mathcal{R}_{nat}^\star:=\inf_f \mathcal{R}_{nat}(f)$, $\mathcal{R}^\star_\phi:=\inf_f \mathcal{R}_\phi(f)$ and $\psi^{-1}$ is the inverse function of the $\psi$-transform of $\phi$.
\end{theorem}

The proof is presented in Appendix \ref{ap:proof}. Theorem \ref{th:final} tells us that our proposed method provides an upper bound on the robust error of the model. To push further, we prove that the suggested loss is tighter than that of TRADES under a weak assumption on the path $\gamma(\cdot)$ in Appendix \ref{ap:proof}.

\section{Experiments} \label{sec:Experiments}
In this section, we describe a set of experiments conducted to verify the advantages of the proposed method. 

\subsection{Experimental setup}
For MNIST, we train LeNet \cite{lecun1998gradient} for 50 epochs with the Adam optimizer. The initial learning rate is 0.001 and it is divided by 10 at 30 and 40 epoch. We use PGD$^{40}$ to generate adversarial examples in the training session with a step-size of $0.02$. No preprocessing or input transformation is used.
For CIFAR10, we train a Wide-ResNet (WRN-28-10) \cite{he2016deep} for 100 epochs using SGD with momentum of 0.9 and weight decay of $5\times10^{-4}$. We use cyclic learning rate schedule \cite{smith2017cyclical}. We use 0.3 as the maximum learning rate and a total of 30 epochs for training. PGD$^{10}$ to generate adversarial examples in the training session with a step-size of $2/255$. Horizontal flip and cropping are used for data augmentation. For both datasets, the robustness regularization hyper-parameter is set to $\beta=5$ for TRADES and the proposed method. We use PyTorch \cite{paszke2019pytorch} and Torchattacks \cite{kim2020torchattacks} for all experiments.
For more additional experiments including the results on CIFAR100 and different model architectures, please refer to Appendix \ref{ap:additional}.
 
\subsection{Reduced negative effect and benefits} \label{subsec:effect}

\begin{figure}
    \centering
    \includegraphics[width=\linewidth]{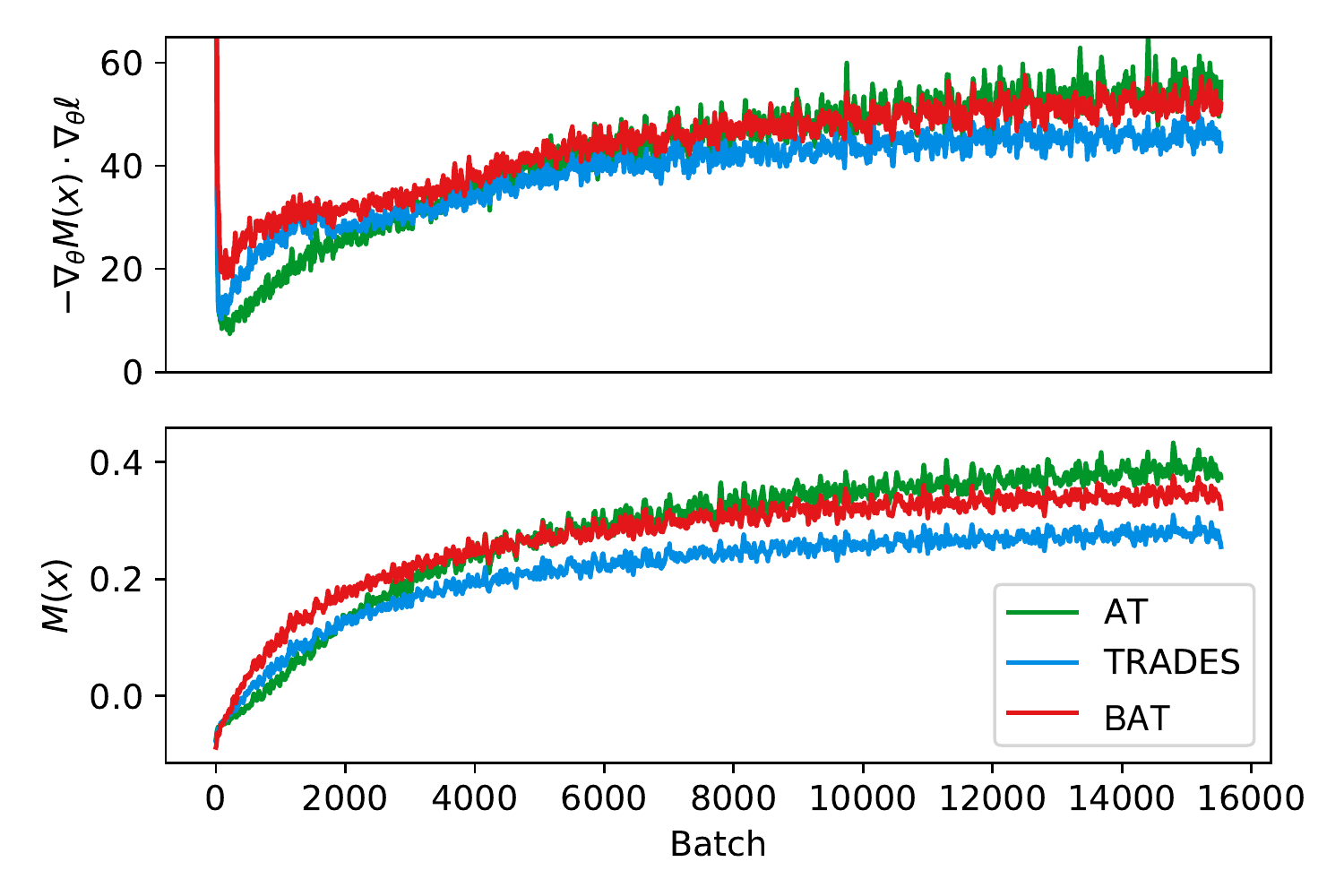}
    \caption{Analysis on the margin during the first 40 epochs. Top: the expected margin increase $- \nabla_\theta M(\vx) \cdot \nabla_\theta \ell$ of each method. Bottom: the actual margin $M(\vx)$ of each method.}
    \label{fig:batch}
\end{figure}

To verify whether the proposed regularizer mitigates the negative effect in \Secref{sec:motivation}, we first observe the effect of the gradient of the loss on the margin, $- \nabla_\vtheta M(\vx) \cdot \nabla_\vtheta \ell$. This indicates the expected margin increase by the weight update with the loss $\ell$. Then, we measure the actual margin $M(\vx)$. \Figref{fig:batch} shows that the proposed method mitigates the negative effect of the regularization term during training. Compared to TRADES, the proposed method shows a higher expected increase in the margin, and this enables the model to learn a large margin.
Thus, by introducing the intermediate probability $\tilde{\vp}$, we successfully encourage the model to reduce the negative effect of the regularization term on maximizing the margin.

The norm of gradient also serves to explain the advantage of the proposed method. As prior works discovered \cite{liu2020loss, dong2021exploring}, a larger gradient norm in the initial training phase enables the model to escapes the suboptimal region. To provide a fair comparison for different training methods, we normalize the norm of gradient by the L2 norm of the loss as follows:
\begin{equation} \label{eq:gradnorm}
\begin{split}
    ||\nabla_\vtheta \hat{\ell}(\vx, \vy)||_2 = \frac{||\nabla_\vtheta \ell(\vx, \vy)||_2}{||\ell(\vx, \vy) ||_2} 
\end{split}
\end{equation}

\begin{figure}
    \centering
    \includegraphics[width=0.9\linewidth]{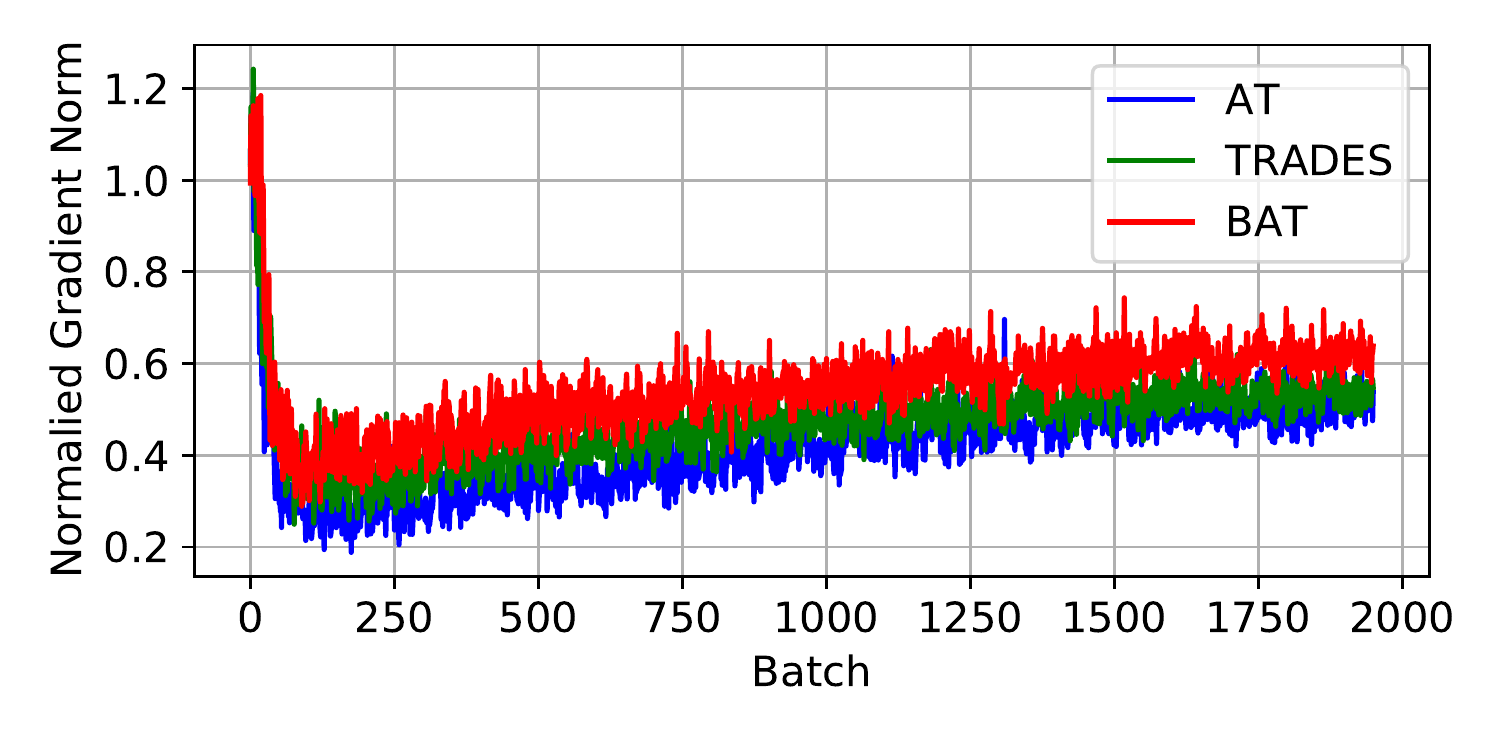}
    \caption{Normalized gradients of each loss term on CIFAR10 during the first 2000 batches. The large gradient magnitude helps quickly escapes the initial suboptimal region \cite{dong2021exploring}.
    \label{fig:gradnorm}
    }
\end{figure}

As shown in \Figref{fig:gradnorm}, the gradients of AT shows the smallest normalized gradient norm among different training methods. This implies that AT has difficulty escaping from initial suboptimal region \cite{liu2020loss}. It is also supported by the experiments for a larger maiximum perturbation in \Figref{fig:mnist_eps}. 
Compared to AT, TRADES shows a higher norm of the gradients. This is consistent to the observation that TRADES provides more gradient stability with the continuous loss landscape \cite{dong2021exploring}. However, TRADES has difficulty reaching the global optima in \Figref{fig:mnist_eps}. This can imply that a higher norm of the gradient is still required. The proposed method shows the highest normalized gradient norm and stands out in having stable convergence even for a larger perturbation conditions in \Secref{sec:Experiments} from the advantage of mitigating the negative effect.

\subsection{Balanced margin and smoothness}

\begin{figure*}
    \centering
    \begin{subfigure}[ht]{.30\textwidth}
      \includegraphics[width=1\columnwidth]{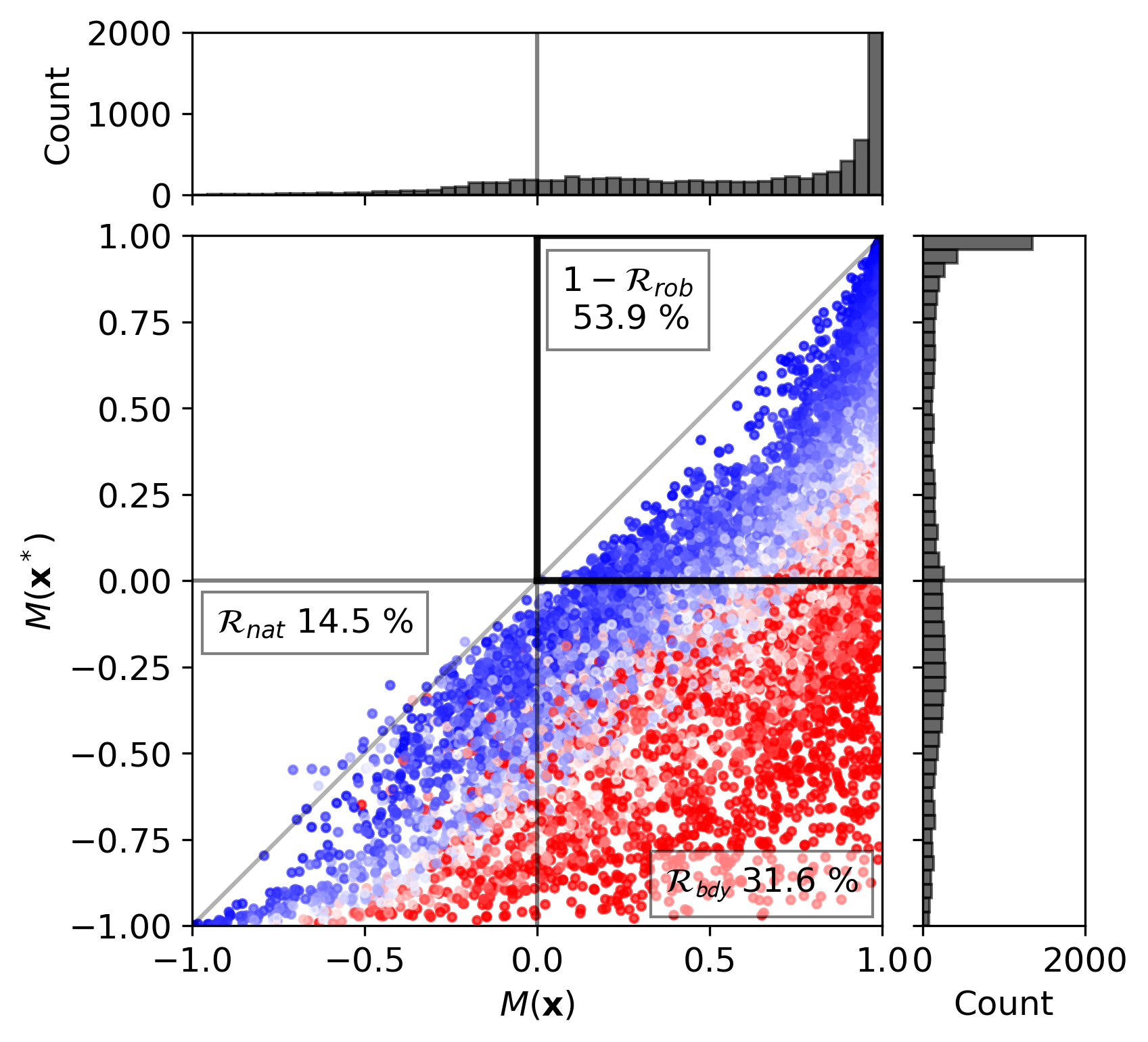}
      \caption{AT}
      \label{fig:quadrant_AT}
    \end{subfigure}
    \begin{subfigure}[ht]{.30\textwidth}
      \includegraphics[width=1\columnwidth]{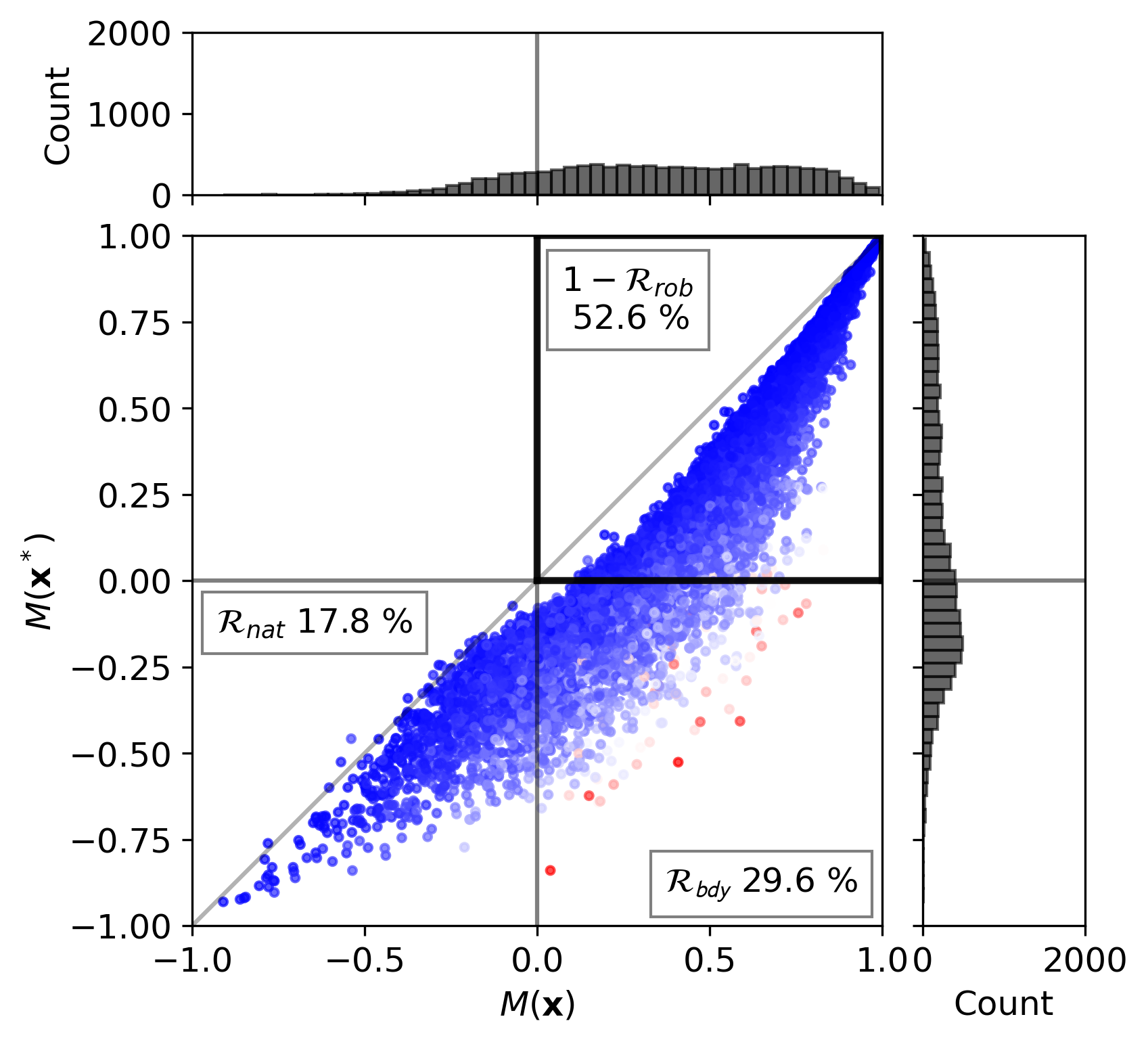}
      \caption{TRADES}
      \label{fig:quadrant_Trades}
    \end{subfigure}
    \begin{subfigure}[ht]{.30\textwidth}
      \includegraphics[width=1\columnwidth]{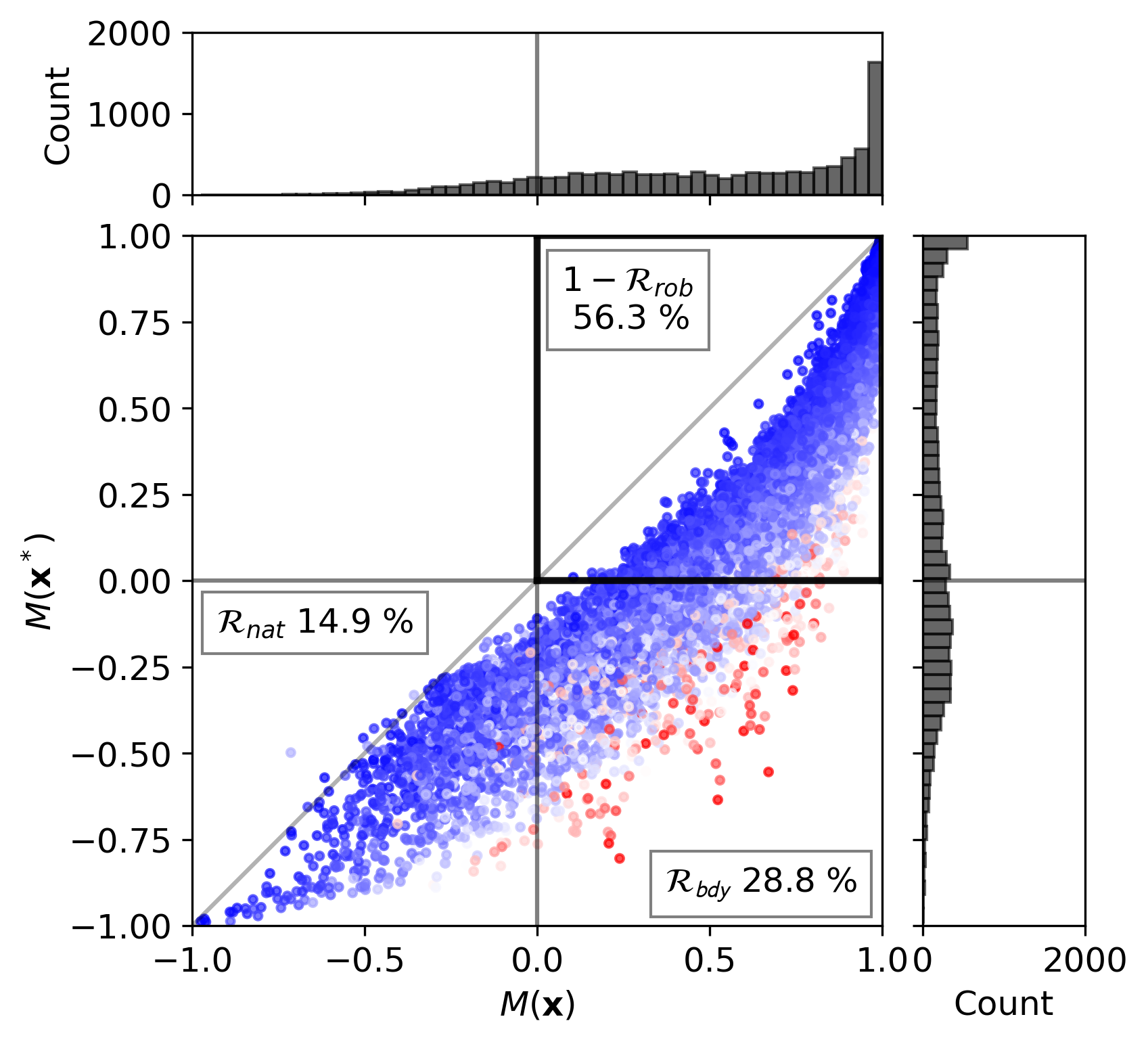}
      \caption{BAT}
      \label{fig:quadrant_ours}
    \end{subfigure}       
    \begin{subfigure}[ht]{.05\textwidth}
      \includegraphics[width=1\columnwidth]{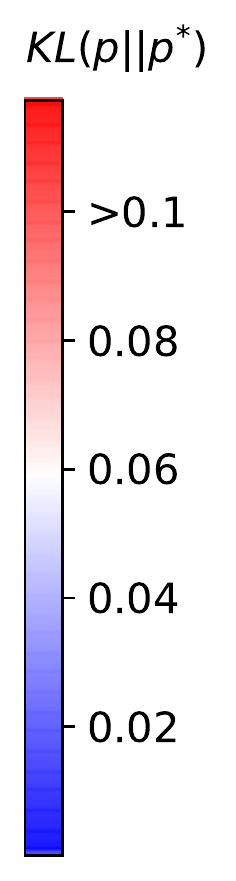}
    \end{subfigure}           
    \caption{Distribution of the margins $M(\vx)$ and $M(\vx^*)$. Each point indicates each test example, and the color of each point indicates the KL divergence loss $\text{KL}(\vp\vert\vert \vp^*)$. The darker red ones indicate a higher KL divergence loss.}
    \label{fig:quadrant}
\end{figure*}

In the previous section, we verified that the proposed method successfully mitigates the negative effect during the initial training phase. Now, we investigate whether the proposed method achieves sufficient smoothness while mitigates the negative effect until the end of training. To illustrate margin and smoothness of the proposed method in more detail, we plot pairs of the margins of clean examples and corresponding adversarial examples with their KL divergence in \Figref{fig:quadrant}. In each plot, the upper and the right histogram shows the distribution of $M(\vx)$ and $M(\vx^*)$, respectively. We generated an adversarial example $\vx^*$ with PGD$^{50}$. Then, we colored each point by the KL divergence $\text{KL}(\vp_\vtheta(\vx)\vert\vert \vp_\vtheta(\vx^*))$ to measure the smoothness. The red points have high KL divergence (poor smoothness) and the blue points have low KL divergence (better smoothness). Note that there is no point in the second quadrant (\textbf{Quadrant \RNum{2}}) because adversarial attacks generally do not make incorrect examples ($M(\vx)<0$) correctly classified ($M(\vx^*)>0$).

In summary, each quadrant corresponds to:
\begin{itemize}
    \item \textbf{Quadrant \RNum{1}:} $M(\vx^*)>0$.
    
    \hfill $\rightarrow$ Adversarial robustness (1 - $\mathcal{R}_{rob}$).
    
    \item \textbf{Quadrant \RNum{3}:} $M(\vx)<0$. 
    
    \hfill $\rightarrow$ Natural classification error ($\mathcal{R}_{nat}$).
    
    \item \textbf{Quadrant \RNum{4}:} $M(\vx)>0$ and $M(\vx^*)<0$. 
    
    \hfill $\rightarrow$ Boundary error ($\mathcal{R}_{bdy}$) by \eqref{eq:decom}.
\end{itemize}

Here, $\mathcal{R}_{nat}(f):=\mathbb{E}_{(\boldsymbol{x},y)} \boldsymbol{1}\{\argmax_i f(\boldsymbol{x})_i \neq y\}$  is the natural classification error and $\mathcal{R}_{rob}(f):=\mathbb{E}_{(\boldsymbol{x},y)} \boldsymbol{1}\{ \exists \boldsymbol{x'}\in\mathbb{B}(\boldsymbol{x},\epsilon) \text{ s.t. } \argmax_i f(\boldsymbol{x'})_i \neq y \}$ is the robust error.
Then, $\mathcal{R}_{rob}(f)$ can be decomposed as follows:
\begin{equation} \label{eq:decom}
\begin{split}
    & \mathcal{R}_{rob}(f) = \mathcal{R}_{nat}(f) + \mathcal{R}_{bdy}(f)
\end{split}
\end{equation}
where $\mathcal{R}_{bdy}(f):=\mathbb{E}_{(\boldsymbol{x},y)} \boldsymbol{1}\{\argmax_i f(\boldsymbol{x})_i = y, \exists \vx'\in\sB(\vx,\epsilon) \text{ s.t. } \argmax_i f(\vx)_i \neq \argmax_i f(\vx')_i\}$ is the boundary error in \cite{zhang2019theoretically}. Thus, the ultimate purpose of adversarial training is to move all the points to the first quadrant.

The results are shown in \Figref{fig:quadrant}. Compared to TRADES, which has only a few samples with a high margin $M(\vx)$ and $M(\vx^*)$, the proposed method shows better margin distributions near 1 for both clean and adversarial examples. This implies that our method successfully mitigates the negative effect of the regularization term on maximizing the margin as discussed in \Secref{sec:method}. As a result, our method achieves a lower natural classification error (14.9\%) than that of TRADES (17.8\%). In addition, the proposed method have more less red points, which implies that the proposed method provide smoothness that AT cannot provide. Due to the increased smoothness, the boundary error of the proposed method ($\mathcal{R}_{bdy}=28.8\%$) is lower than that of AT ($\mathcal{R}_{bdy}=31.6\%$). In summary, the proposed method provide the balanced margin and smoothness with better robustness.

\subsection{Robustness} \label{subsec:whitebox}
In this section, we verify the robustness of the proposed method. In addition to AT and TRADES, we consider MART \cite{wang2019improving}, which aims to maximize the margin and recently achieved the best performance by focusing on misclassified examples. Here, we use PGD$^{50}$ with 10 random restarts. The step-size is fixed to $0.01$ and $2/255$ for MNIST and CIFAR10, respectively. Furthermore, we also consider AutoAttack \cite{croce2020reliable}, which is a combination of three white-box attacks \cite{croce2020reliable, croce2019minimally} and one black-box attack \cite{andriushchenko2019square}. Note that AutoAttack is by far the most reliable attack to measure robustness. Each experiment was repeated over 3 runs with different random seeds.
 
\begin{table}[t]
\caption{Robustness accuracy (\%) on MNIST.}
\vskip 0.15in
\label{table:whitebox_summary_mnist}
\begin{tabular}{lccg}
\multicolumn{1}{l|}{Method}            & \multicolumn{1}{c}{Clean} & \multicolumn{1}{c}{PGD$^{50}$} & \multicolumn{1}{g}{AutoAttack} \\ \hline

\multicolumn{4}{l}{}                                                                                                                                  \\
\multicolumn{4}{l}{\textbf{MNIST ($\epsilon=0.3$)}}                                                                                               \\ \hline
\multicolumn{1}{l|}{AT}            & 98.79$\pm$0.23                     & 91.69$\pm$1.25                     & 88.64$\pm$0.61                     \\ 

\multicolumn{1}{l|}{TRADES}         & \textbf{98.89$\pm$0.01}                     & 93.70$\pm$0.01                     & \textbf{92.31$\pm$0.21}                     \\ 

\multicolumn{1}{l|}{MART$^\dagger$}           & 98.78$\pm$0.16                     & 92.37$\pm$1.21                     & 89.62$\pm$1.44                     \\ 

\multicolumn{1}{l|}{BAT}           & 98.79$\pm$0.06                     & \textbf{93.97$\pm$0.16}                     & 92.19$\pm$0.13                     \\ 

\multicolumn{4}{l}{}                                                                                                                                  \\
\multicolumn{4}{l}{\textbf{MNIST ($\epsilon=0.45$)}}                                                                                               \\ \hline
\multicolumn{1}{l|}{AT}            & 11.35$\pm$0.00                     & 11.35$\pm$0.00                     & 11.35$\pm$0.00                     \\

\multicolumn{1}{l|}{TRADES}         & \textbf{99.42$\pm$0.05}                     & 11.34$\pm$0.01                     & 0.53$\pm$0.26                     \\

\multicolumn{1}{l|}{MART$^\dagger$}           & 13.13$\pm$2.54                     & 7.80$\pm$3.73                     & 0.93$\pm$1.32                     \\ 

\multicolumn{1}{l|}{BAT}            & 97.72$\pm$0.26                     & \textbf{88.20$\pm$0.57}                     & \textbf{76.09$\pm$1.65}                     \\ 
\end{tabular}
\small{\raggedright \\ $^\dagger$ SGD with the initial learning rate 0.01 is used following \citet{wang2019improving} because MART converges to a constant function with Adam. \par}
\end{table}

\begin{table}
\caption{Robustness accuracy (\%) on CIFAR10.}
\label{table:whitebox_cifar10_c}
\centering
\begin{tabular}{lccg}
\multicolumn{1}{l|}{Method} & \multicolumn{1}{c}{Clean}                
& \multicolumn{1}{c}{PGD$^{50}$}                    
& \multicolumn{1}{g}{AutoAttack}                      \\ \hline
\multicolumn{4}{l}{}                                                                                                                                                                              \\
\multicolumn{4}{l}{\textbf{CIFAR10 ($\epsilon=8/255$)}}                                                                                                                                           \\ \hline
\multicolumn{1}{l|}{AT}               & \textbf{85.65$\pm$0.33}                     & 53.64$\pm$0.03                     & 50.87$\pm$0.22                     \\ 

\multicolumn{1}{l|}{TRADES}            & 82.22$\pm$0.12                     & 52.14$\pm$0.08                     & 48.90$\pm$0.35                     \\ 

\multicolumn{1}{l|}{MART}              & 77.51$\pm$0.46                     & 53.87$\pm$0.08                     & 48.25$\pm$0.06                     \\ 

\multicolumn{1}{l|}{BAT}               & 84.84$\pm$0.28                     & \textbf{55.64$\pm$0.37}                     & \textbf{52.41$\pm$0.02}                     \\ 

\multicolumn{4}{l}{}                                                                                                                                                                              \\
\multicolumn{4}{l}{\textbf{CIFAR10 ($\epsilon=16/255$)}}                                                                                                                                          \\ \hline
\multicolumn{1}{l|}{AT}    & 72.43$\pm$0.01                                       & 29.01$\pm$0.13                   & 24.24$\pm$0.51                    \\
\multicolumn{1}{l|}{TRADES} & 70.01$\pm$0.44                                       & 24.52$\pm$0.06                  & 14.63$\pm$0.22                     \\
\multicolumn{1}{l|}{MART}    & 65.97$\pm$0.54                              & \textbf{32.65$\pm$0.40}        & 23.23$\pm$0.14                   \\
\multicolumn{1}{l|}{BAT}    & \textbf{77.56$\pm$0.01}                              & 30.79$\pm$0.35         & \textbf{25.06$\pm$0.37}    
\end{tabular}
\end{table}

As shown in \Tabref{table:whitebox_summary_mnist}, for MNIST with $\epsilon=0.3$, all defenses show high robustness.
However, for a large $\epsilon=0.45$, all comparison methods converge to a constant function or fail to gain robustness. 
In other words, the existing methods have difficulty converging to the global optimal. For the cases of AT and MART, they have the term that maximizes the margin of adversarial examples so that it can have difficulty in convergence \cite{dong2021exploring}. In contrast, TRADES also fails to achieve stable robustness, because a larger perturbation brings stronger negative effect of $\text{KL}(\vp||\vp^*)$ as we discussed in \Secref{sec:motivation}. However, the proposed method shows stable results even for $\epsilon=0.45$. Considering that the difference between TRADES and the proposed method is that the usage of bridging, this result tells us that the convergence becomes much more easier by using the proposed bridged loss.

The proposed method also shows the best robustness on CIFAR10 (\Tabref{table:whitebox_cifar10_c}). Specifically, for $\epsilon=16/255$, the proposed method achieves 77.56\% of standard accuracy with is 5\% higher than AT. Compared to TRADES and MART, it is 6\% and 12\% higher, respectively. Simultaneously, it also achieves the best robustness 25.06\% against AutoAttack. Note that the robustness of TRADES is only 14.63\%, which shows the weakness of TRADES for a largeR perturbation.

\paragraph{Robust self-training.}
Recently, it has been found that using additional unlabeled data can greatly improve the standard accuracy and robustness \cite{carmon2019unlabeled, uesato2019labels, zhai2019adversarially, najafi2019robustness}. Thus, following \citet{carmon2019unlabeled}, we use the additional data (500K images) and the cosine learning rate annealing \cite{loshchilov2016sgdr} without restarts for 200 epochs.

\Tabref{table:semi} shows the results of the experiment with unlabeled data. For $\epsilon=8/255$, the proposed method shows the best robustness against AutoAttack. Especially, for $\epsilon=16/255$, the proposed method outperforms the other methods by a large margin. In particular, compared to TRADES, the proposed method shows an approximately 6\% improvement in the standard accuracy, while the robustness is also greatly increased (3\%).

\begin{table}[t]
\caption{Robustness accuracy (\%) on CIFAR10 with unlabeled data.} \label{table:semi}
\vskip 0.15in
\centering
\begin{tabular}{lccg}
\multicolumn{1}{l|}{Method}            & \multicolumn{1}{c}{Clean} & 
\multicolumn{1}{c}{PGD$^{50}$} & \multicolumn{1}{g}{AutoAttack} \\ \hline
\multicolumn{4}{l}{}                                                                                                                                                                              \\
\multicolumn{4}{l}{\textbf{CIFAR10 ($\epsilon=8/255$)}}                                                                                                                                           \\ \hline
\multicolumn{1}{l|}{AT-RST}               & \textbf{91.53}
                    & 59.89
                   & 58.41
                       \\
\multicolumn{1}{l|}{TRADES-RST} & 89.73
                    & 61.87
                   & 59.45
                       \\
\multicolumn{1}{l|}{MART-RST}               & 89.71
                    & 62.11
                   & 57.97
                       \\
\multicolumn{1}{l|}{BAT-RST}               & 89.61
                    & \textbf{62.38}
                   & \textbf{59.54}
                       \\
\multicolumn{4}{l}{}                                                                                                                                                                              \\
\multicolumn{4}{l}{\textbf{CIFAR10 ($\epsilon=16/255$)}}                                                                                                                                       \\ \hline                                                            
\multicolumn{1}{l|}{AT-RST}               & 83.36
                    & 29.48
                   & 25.54
                       \\
\multicolumn{1}{l|}{TRADES-RST} & 78.20
                    & \textbf{34.14}
                   & 24.96
                       \\
\multicolumn{1}{l|}{MART-RST}               & 81.24
                    & 33.17
                   & 25.77
                       \\
\multicolumn{1}{l|}{BAT-RST}               & \textbf{84.07}
                    & 33.78
                   & \textbf{27.70}
                       \\

\end{tabular}
\end{table}

\section{Conclusion}
In this paper, we investigated the existing adversarial training methods from the perspective of margin and smoothness of the network.
We found that AT and TRADES have different characteristics in terms of margin and smoothness due to their different regularizers. We mathematically proved that the regularization term designed for smoothness has a negative effect on training a larger margin.
To this end, we proposed a new method that mitigates the negative effect by bridging the gap between clean and adversarial examples and achieved stable and better performance. 
Our investigation on margin and smoothness can provide a new perspective to better understand the adversarial robustness and to design a robust model.

\section*{Acknowledgment}
This work was supported by the National Research Foundation of Korea (NRF) grant funded by the Korean government (MSIT) (NRF-2019R1A2C2002358).

\bibliography{ms}
\bibliographystyle{icml2021}

\newpage
\onecolumn
\appendix
\counterwithin{figure}{section}
\counterwithin{table}{section}
\counterwithin{equation}{section}

\section{Proofs of theoretical results} \label{ap:proof}

\subsection{Proof of Proposition \ref{lem:gradient}}

\begin{varlem}{lem:gradient}
Let $\vp=\vp_\vtheta(\vx)$, $\vp^*=\vp_\vtheta(\vx^*)$, $\nabla=\nabla_\vtheta$, and $t=\max_{i\neq y}\vp$. If $\log\alpha_y>0$ and $\log\alpha_t <0$ for $\log \alpha_i =\log\frac{p_i}{p^*_i}$, then the gradient descent direction of 
$\text{KL}(\vp||\vp^*)$ is aligned with the gradient direction that minimizes the margin $M(\vx)=p_y-p_t$ by penalizing $p_i$ with the scale of $\log \alpha_i$.
\begin{align*}
    -\nabla \text{KL}(\vp||\vp^*) = - (\nabla p_y)^T \log\alpha_y - (\nabla p_t)^T \log\alpha_t + c
\end{align*}
where $c$ is a linear combination of other gradient directions.
\end{varlem}
\begin{proof} 
\begin{align*}\label{eq:alpha}
    &\nabla \text{KL}(\vp||\vp^*) 
    \\ = &\nabla\left(\vp^T\log \vp-\vp^T\log \vp^*\right)
    \\ = & (\nabla \vp)^T \log \vp + (\nabla \vp)^T \1 - (\nabla \vp)^T \log \vp^* - (\nabla \vp^*) ^T \frac{\vp}{\vp^*}
    \\ = & (\nabla \vp)^T \log \frac{\vp}{\vp^*} - (\nabla \vp^*) ^T \frac{\vp}{\vp^*}
    \\ = & (\nabla \vp)^T \log \valpha - (\nabla \vp^*) ^T \valpha. \numberthis
\end{align*}
\end{proof}

\subsection{Proof of Theorem \ref{th:final}}
To give a self-contained overview, we follow \cite{bartlett2006convexity}. In the binary classification case, given a sample $\boldsymbol{x} \in \mathcal{X}$ and a label $y \in \{-1, 1\}$, a model can be denoted as $f:\mathcal{X} \rightarrow \mathbb{R}$. We use $\text{sign}(f(\boldsymbol{x}))$ as a prediction value of $y$. Given a surrogate loss $\phi$, the conditional $\phi$-risk for $\eta \in [0,1]$ can be denoted as $H(\eta):=\inf_{\alpha\in\mathbb{R}}(\eta \phi (\alpha) + (1-\eta)\phi(-\alpha))$. Similarly, we can define $H^-(\eta):=\inf_{\alpha(2\eta-1)\leq 0 } (\eta \phi (\alpha) + (1-\eta)\phi(-\alpha))$. Now, we assume the surrogate loss $\phi$ is classification-calibrated, so that $H^-(\eta) > H(\eta)$ for any $\eta \neq 1/2$. Then, the $\psi$-transform of a loss function $\phi$, which is the convexified version of $\hat{\psi}(\theta)=H^-(\frac{1+\theta}{2})-H(\frac{1+\theta}{2})$, is continuous convex function on $\theta \in [-1,1]$.

In the adversarial training framework, we train a model to reduce the robust error $\mathcal{R}_{rob}(f):=\mathbb{E}_{(\boldsymbol{x},y)} \boldsymbol{1}\{ \exists \boldsymbol{x'}\in\mathbb{B}(\boldsymbol{x},\epsilon) \text{ s.t. } f(\boldsymbol{x'})y \leq 0 \}$ where $\mathbb{B}(\boldsymbol{x},\epsilon)$ is a ball around an example $x$ with a maximum perturbation $\epsilon$.
Here, $\boldsymbol{1}\{C\}$ denotes an indicator function which outputs 1 if the condition $C$ is true and 0 otherwise. As \citet{zhang2019theoretically} proposed, $\mathcal{R}_{rob}(f)$ can be decomposed as follows:
\begin{equation} \label{eq:decom_2}
\begin{split}
    & \mathcal{R}_{rob}(f) = \mathcal{R}_{nat}(f) + \mathcal{R}_{bdy}(f)
\end{split}
\end{equation}
where the natural classification error $\mathcal{R}_{nat}(f):=\mathbb{E}_{(\boldsymbol{x},y)} \boldsymbol{1}\{f(\boldsymbol{x})y\leq0\}$ and boundary error  $\mathcal{R}_{bdy}(f):=\mathbb{E}_{(\boldsymbol{x},y)} \boldsymbol{1}\{f(\boldsymbol{x})y>0, \exists \vx'\in\sB(\vx,\epsilon) \text{ s.t. } f(\vx)f(\vx')\leq0\}$.
 By definition, following inequality is satisfied:
\begin{equation} \label{eq:bdy}
\begin{split}
    \mathcal{R}_{bdy}(f) &= \mathbb{E}_{(\boldsymbol{x},y)} \boldsymbol{1}\{\boldsymbol{x} \in \mathbb{B}(\text{DB}(f), \epsilon), f(\boldsymbol{x})y>0\}
    \\&\leq \mathbb{E}_{(\boldsymbol{x},y)} \boldsymbol{1}\{\boldsymbol{x} \in \mathbb{B}(\text{DB}(f), \epsilon)\}
    \\&= \mathbb{E} \max_{\boldsymbol{x'} \in \mathbb{B}(\boldsymbol{x}, \epsilon)} \boldsymbol{1}\{f(\boldsymbol{x'})\neq f(\boldsymbol{x})\}
    \\&= \mathbb{E} \max_{\boldsymbol{x'} \in \mathbb{B}(\boldsymbol{x}, \epsilon)} \boldsymbol{1}\{\beta f(\boldsymbol{x'})f(\boldsymbol{x})<0\}.
\end{split}
\end{equation}

Let, $\mathcal{R}_{nat}^\star:=\inf_f \mathcal{R}_{nat}(f)$ and $\mathcal{R}^\star_\phi:=\inf_f \mathcal{R}_\phi(f)$ where $\mathcal{R}_{\phi}(f):= \mathbb{E}_{(\boldsymbol{x},y)} \phi\{f(\boldsymbol{x})y\leq0\}$ is a surrogate loss with a surrogate loss function $\phi$. Then, under Assumption \ref{asu:gamma}, following inequality is satisfied by \eqref{eq:decom_2} and \eqref{eq:bdy}.
\begin{equation} \label{eq:ineq}
\begin{split}
    \mathcal{R}_{rob}(f) - \mathcal{R}_{nat}^\star \leq &  \psi^{-1}(\mathcal{R}_{\phi}(f)-\mathcal{R}^\star_\phi)
    + \mathbb{E} \max_{\boldsymbol{x'} \in \mathbb{B}(\boldsymbol{x}, \epsilon)} \boldsymbol{1}\{\beta f(\boldsymbol{x'})f(\boldsymbol{x})<0\}
\end{split}
\end{equation}

We push further the analysis by considering generalized intermediate value theorem.
\begin{theorem}[Generalized intermediate value theorem]
\label{th:givt}
Let $f:\mathcal{X}\subset\mathbb{R}^d \rightarrow \mathbb{R}$ be a continuous map and a continuous path $\gamma:[0,1]\rightarrow\mathcal{X}$ such that $\gamma(0)=\boldsymbol{a}$ and $\gamma(1)=\boldsymbol{b}$. If $f(\boldsymbol{a})f(\boldsymbol{b}) < 0$, then there is at least one $u\in(0,1)$ that satisfies $f(\gamma(u))=0$.
\end{theorem}

\begin{proof}
Consider $g(\cdot)=f(\gamma(\cdot))$. Then, $g:[0,1]\rightarrow\mathbb{R}$ is a continuous function with $g(0)=f(\boldsymbol{a})$ and $g(1)=f(\boldsymbol{b})$. Since $g(0)g(1)<0$, by the intermediate value theorem, there exists $u\in(0,1)$ such that $g(u)=0$. 
\end{proof}

By Theorem \ref{th:givt}, given example $\vx$, adversarial example $\boldsymbol{x}^*$ and a continuous path $\gamma(\cdot)$ such that $\gamma(0)=\boldsymbol{x}$ and $\gamma(1)=\boldsymbol{x}^*$, following inequality is satisfied:
\begin{equation} \label{eq:newineq}
\begin{split}
    &\boldsymbol{1}\{\beta f(\boldsymbol{x'})f(\boldsymbol{x})<0\} \leq \sum_{k=0}^{m-1}\boldsymbol{1}\{\beta f(\gamma(\frac{k}{m}))f(\gamma(\frac{k+1}{m}))<0\}
\end{split}
\end{equation}
where $m$ is a hyper-parameter for dividing the path $\gamma(\cdot)$.
Now, we establish a new upper bound on $\mathcal{R}_{rob}(f) - \mathcal{R}_{nat}^\star$.

\begin{varthm}{th:final}
Given a sample $\vx$ and a positive $\beta$, let $\gamma:[0,1]\rightarrow\mathcal{X}$ be a continuous path from $\gamma(0)=\boldsymbol{x}$ to $\gamma(1)=\boldsymbol{x}^*$ where $\boldsymbol{x}^* = \arg\max_{\boldsymbol{x'} \in \mathbb{B}(\boldsymbol{x}, \epsilon)} \boldsymbol{1}\{\beta f(\boldsymbol{x'})f(\boldsymbol{x})<0\}$. Then, we have
\begin{align*}
    \mathcal{R}_{rob}(f) - \mathcal{R}_{nat}^\star & \leq  \psi^{-1}(\mathcal{R}_{\phi}(f)-\mathcal{R}^\star_\phi)
    + \mathbb{E}_{(\boldsymbol{x},y)}\sum_{k=0}^{m-1}\phi(\beta f(\gamma(\frac{k}{m}))f(\gamma(\frac{k+1}{m}))) 
\end{align*}
where $\mathcal{R}_{nat}^\star:=\inf_f \mathcal{R}_{nat}(f)$, $\mathcal{R}^\star_\phi:=\inf_f \mathcal{R}_\phi(f)$ and $\psi^{-1}$ is the inverse function of the $\psi$-transform of $\phi$.
\end{varthm}

\begin{proof} By \eqref{eq:ineq}, the second inequality holds. Similarly, the third inequality holds by \eqref{eq:newineq}, and the last inequality holds because we choose a classification-calibrated loss $\phi$.
\begin{align*}
    \mathcal{R}_{rob}(f) - \mathcal{R}_{nat}^\star 
    & \leq \psi^{-1}(\mathcal{R}_{\phi}(f)-\mathcal{R}^\star_\phi)
    +  \mathbb{E}_{(\boldsymbol{x},y)}\boldsymbol{1}\{\beta f(\boldsymbol{x'})f(\boldsymbol{x})<0\} 
    \\& \leq \psi^{-1}(\mathcal{R}_{\phi}(f)-\mathcal{R}^\star_\phi)
     +  \mathbb{E}_{(\boldsymbol{x},y)}\sum_{k=0}^{m-1}\boldsymbol{1}\{\beta f(\gamma(\frac{k}{m}))f(\gamma(\frac{k+1}{m}))<0\}
    \\& \leq \psi^{-1}(\mathcal{R}_{\phi}(f)-\mathcal{R}^\star_\phi)
     +  \mathbb{E}_{(\boldsymbol{x},y)}\sum_{k=0}^{m-1}\phi(\beta f(\gamma(\frac{k}{m}))f(\gamma(\frac{k+1}{m})))
\end{align*}
\end{proof}


Following \cite{zhang2019theoretically}, we use KL divergence loss ($\text{KL}$) as a classification-calibrated loss. To do this, we define $p(x):=\sigma(f(x))$ where $\sigma$ is a sigmoid function. Then, a model output with softmax can be denoted as $\vp(x):=[p(x), 1-p(x)]$. In this setting, we can prove that the suggest loss is tighter than that of TRADES under a weak assumption on $\gamma(\cdot)$.

\begin{asu} \label{asu:gamma} $[\vp(\gamma(t))]_y$ is a decreasing function of $t\in[0,1]$, where $[\vp(\cdot)]_y$ indicates the probability corresponding to the correct label $y$.
\end{asu}

\begin{theorem}\label{th:kl}
Under Assumption \ref{asu:gamma}, the KL divergence loss has the following property:
\begin{align*}
    \sum_{k=0}^{m-1}\text{KL}(\vp(\gamma(\frac{k}{m}))||\vp(\gamma(\frac{k+1}{m})) \leq \text{KL}(\vp(\gamma(0))||\vp(\gamma(1))). 
\end{align*}
\end{theorem}

\begin{proof} Let $p_1(x)$, $p_2(x)$, and $p_3(x)$ denotes three different distribution with possible outcomes $x=\{-1, 1\}$ and $0 < p_1(x=1) \leq p_2(x=1) \leq p_3(x=1) < 1$. Then,
\begin{align*}
    &\text{KL}(p_1\vert\vert p_2) + \text{KL}(p_2\vert\vert p_3) - \text{KL}(p_1\vert\vert p_3)
    \\ = & \sum_{x\in\{0,1\}} p_1(x) \ln \frac{p_1(x)}{p_2(x)} + \sum_{x\in\{0,1\}} p_2(x) \ln \frac{p_2(x)}{p_3(x)} - \sum_{x\in\{0,1\}} p_1(x) \ln \frac{p_1(x)}{p_3(x)}
    \\ = & \sum_{x\in\{0,1\}} (p_2(x)-p_1(x)) \ln p_2(x) + \sum_{x\in\{0,1\}} (p_1(x)-p_2(x)) \ln p_3(x)
    \\ = & -\sum_{x\in\{0,1\}} (p_1(x)-p_2(x)) (\ln p_2(x) - \ln p_3(x))
    \\ \leq & \text{ } 0
\end{align*}
The last inequality holds because $p_1(x)-p_2(x)$ and $p_2(x)-p_3(x)$ have the same sign regardless of $x$. Likewise, for $0 < p_1(x=-1) \leq p_2(x=-1) \leq p_3(x=-1) < 1$, the statement also holds true. Thus, by mathematical induction, $\sum_{k=0}^{m-1}\text{KL}(\vp(\gamma(\frac{k}{m}))||\vp(\gamma(\frac{k+1}{m})) \leq \text{KL}(\vp(\gamma(0))||\vp(\gamma(1)))$ under Assumption \ref{asu:gamma}.
\end{proof}

For the multi-class problem, we can extend Theorem \ref{th:kl} by assuming $[\vp(\gamma(u))]_i$ as a monotonic function for each individual component $i\in \mathcal{Y}$.

\section{Additional experiments} \label{ap:additional}

\subsection{Achieving both good margin and smoothness on MNIST}

\begin{figure}
    \centering
    \begin{subfigure}[ht]{.3\textwidth}
      \includegraphics[width=1\columnwidth]{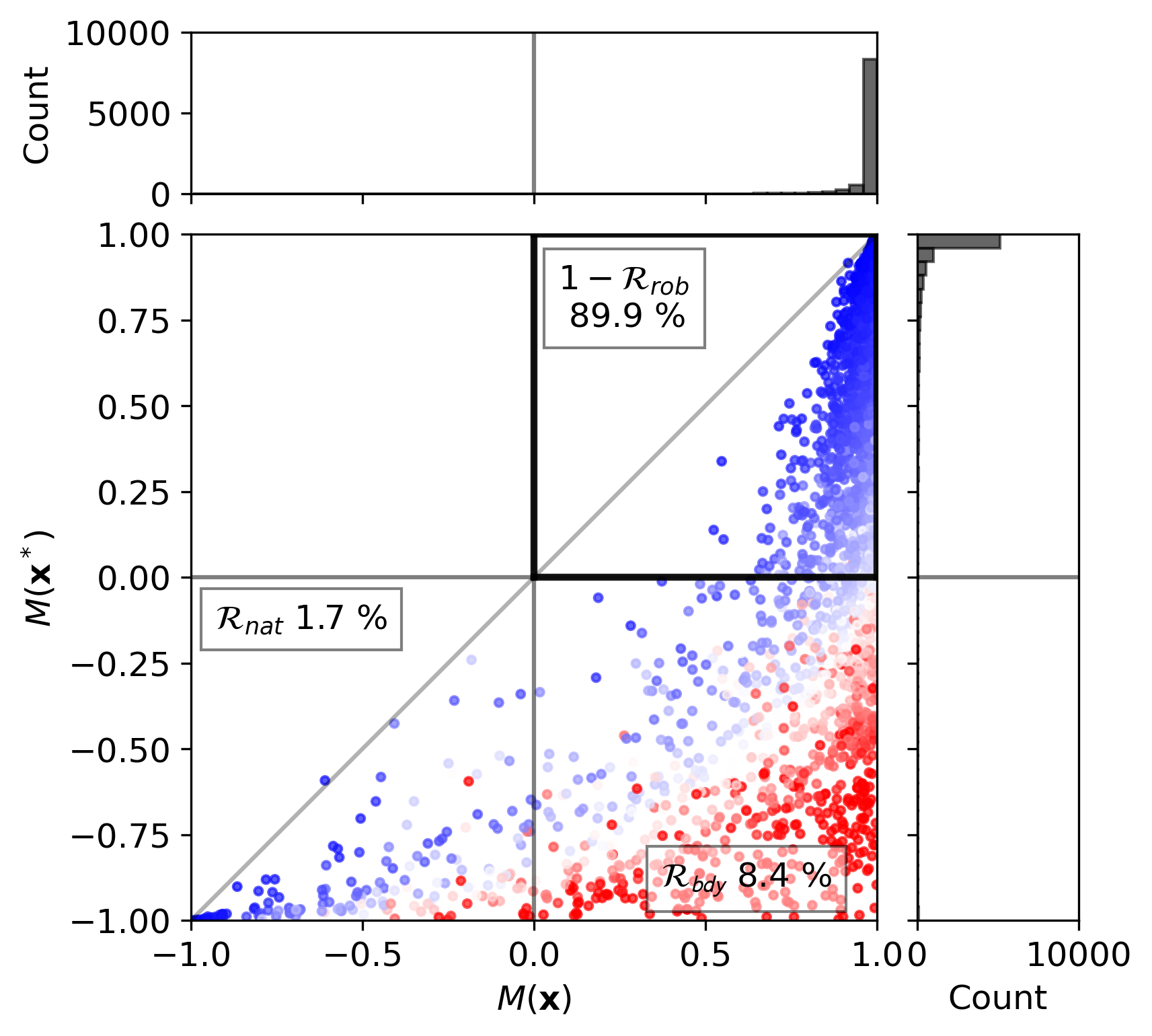}
      \caption{AT}
      \label{fig:mnist_quadrant_AT}
    \end{subfigure}
    \begin{subfigure}[ht]{.3\textwidth}
      \includegraphics[width=1\columnwidth]{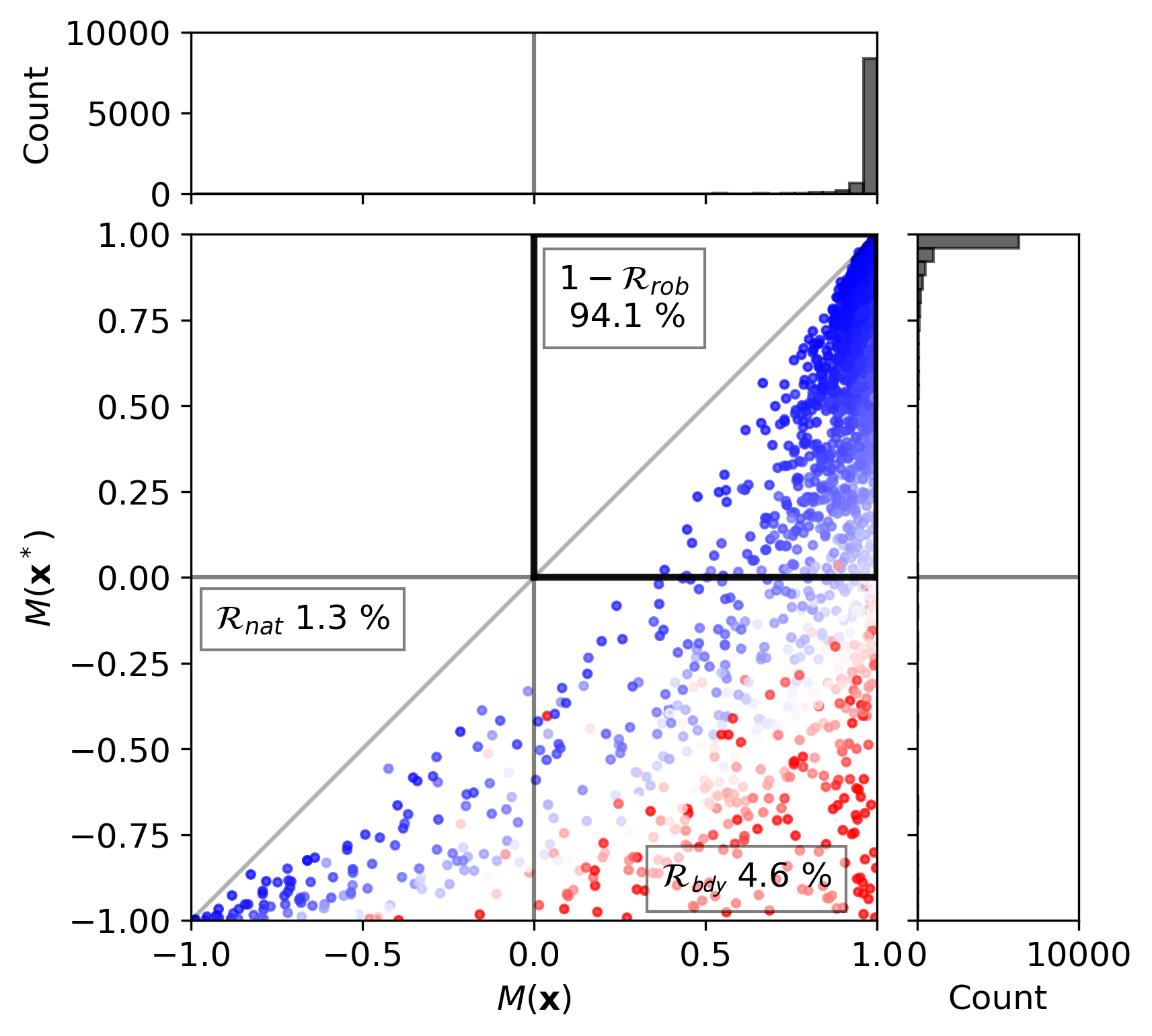}
      \caption{TRADES}
      \label{fig:mnist_quadrant_Trades}
    \end{subfigure}
    \begin{subfigure}[ht]{.3\textwidth}
      \includegraphics[width=1\columnwidth]{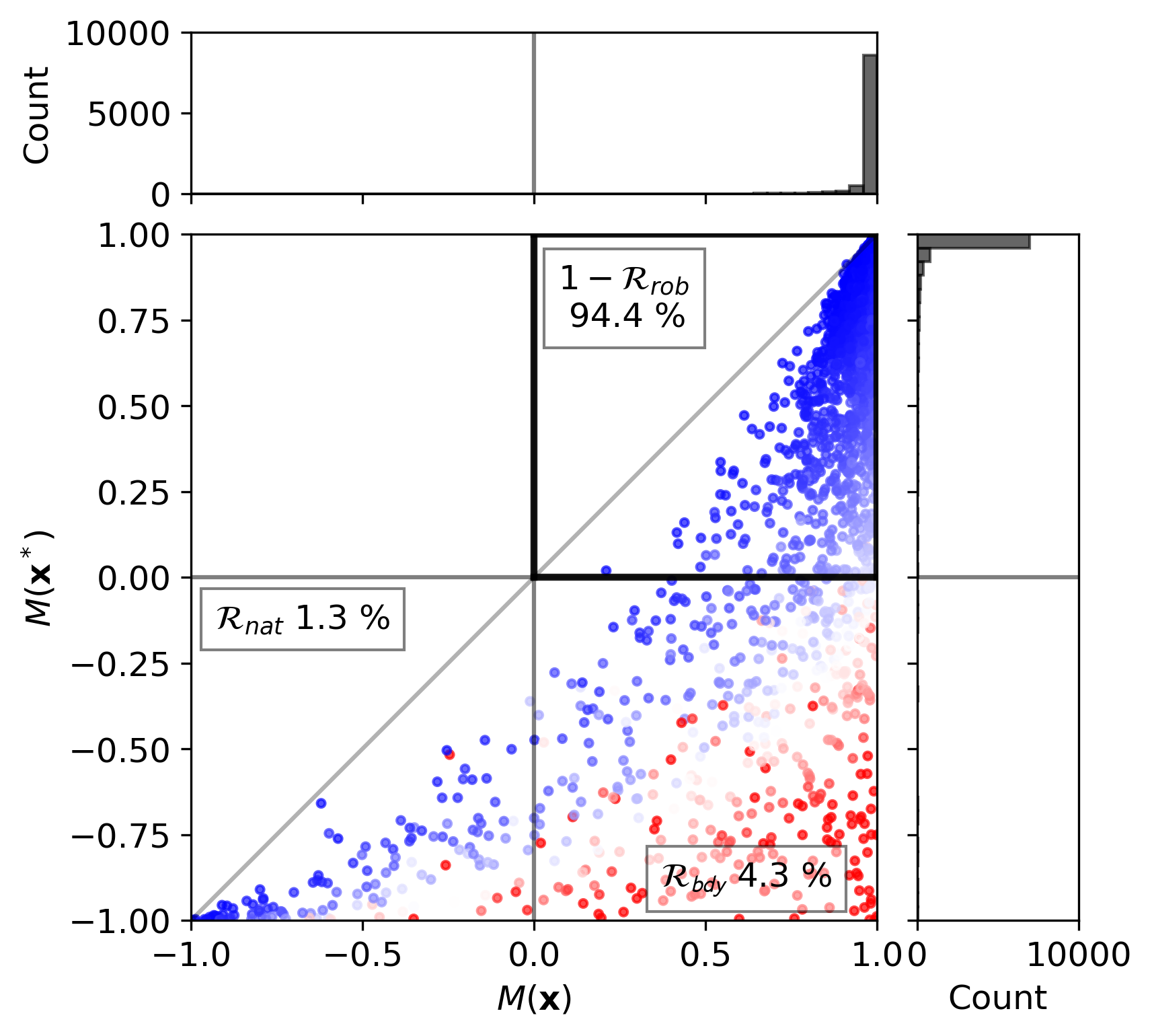}
      \caption{BAT}
      \label{fig:mnist_quadrant_ours}
    \end{subfigure}       
    \begin{subfigure}[ht]{.06\textwidth}
      \includegraphics[width=1\columnwidth]{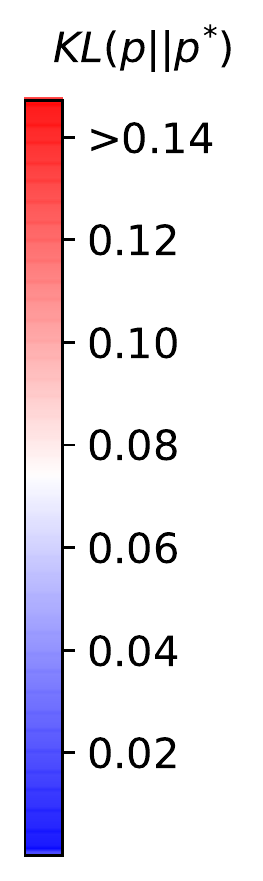}
    \end{subfigure}           
    \caption{Distribution of the margins $M(\vx)$ and $M(\vx^*)$ on MNIST. Each point indicates each test example, and the color of each point indicates the KL divergence loss $\text{KL}(\vp\vert\vert \vp^*)$. The darker red ones indicate a higher KL divergence loss.}
    \label{fig:mnist_quadrant}
\end{figure}

As \Figref{fig:quadrant}, we check the quadrant plots for each method on MNIST. We note that for $\epsilon=0.3$, all methods output near 98.8\% of the standard accuracy so that it is hard to distinguish. Thus, we perform the experiment on $\epsilon=0.35$ which is the maximum perturbation that AT does not converge to a constant function. All the other settings are remained the same. The result is demonstrated in \Figref{fig:mnist_quadrant}. Here again, AT shows a less smoothness than TRADES and the proposed method. In addition, it also shows the highest boundary error $\mathcal{R}_{bdy}$, 8.4\%. The proposed method shows the highest robust accuracy 94.4\% and the lowest boundary error 4.3\%. Compared to TRADES, the proposed method shows better margin. Specifically, the ratio of clean examples with the margin $M(\vx)$ over 0.9 is 91.88\% which is higher than that of TRADES (91.71\%). Moreover, the proposed method also shows higher ratio of adversarial examples with the margin (79.71\%) than that of TRADES (75.94\%).


\subsection{Experiment on CIFAR10}
\paragraph{Step-wise learning rate decay.}

\begin{table}
\caption{Robustness accuracy (\%) on CIFAR10. Each top line indicates the performance of the model at the end of training, and each bottom line (ES) indicates the performance of the best checkpoint by using early stopping.}
\label{table:whitebox_cifar10}
\centering
\begin{tabular}{lcccg}
\multicolumn{1}{l|}{Method}            & \multicolumn{1}{c}{Clean} & \multicolumn{1}{c}{FGSM} & \multicolumn{1}{c}{PGD$^{50}$} & \multicolumn{1}{g}{AutoAttack} \\ \hline

\multicolumn{5}{l}{}                                                                                                                                  \\
\multicolumn{5}{l}{\textbf{CIFAR10 ($\epsilon=8/255$)}}                                                                                               \\ \hline
\multicolumn{1}{l|}{AT}               & 86.85$\pm$0.14                     & 57.33$\pm$0.23                     & 45.33$\pm$0.49                     & 44.61$\pm$0.45                     \\

\multicolumn{1}{r|}{(ES)}           & 84.59$\pm$0.01                     & 59.19$\pm$0.45                     & 53.09$\pm$0.13                     & 50.39$\pm$0.06                     \\

\multicolumn{1}{l|}{TRADES}         & 86.37$\pm$0.36                     & 61.24$\pm$0.30                     & 51.81$\pm$0.04                     & 50.05$\pm$0.03                     \\

\multicolumn{1}{r|}{(ES)}           & 82.85$\pm$0.71                     & 57.73$\pm$0.35                     & 52.24$\pm$0.59                     & 49.25$\pm$0.37                     \\

\multicolumn{1}{l|}{MART}               & 84.79$\pm$0.11                     & 59.81$\pm$0.73                     & 49.91$\pm$0.36                     & 46.16$\pm$0.08                     \\

\multicolumn{1}{r|}{(ES)}           & 78.02$\pm$0.99                     & 57.19$\pm$0.73                     & 53.45$\pm$0.14                     & 48.19$\pm$0.28                     \\

\multicolumn{1}{l|}{BAT}               & 85.36$\pm$0.01                     & 57.53$\pm$0.41                     & 47.64$\pm$0.20                     & 46.28$\pm$0.28                     \\ 

\multicolumn{1}{r|}{(ES)}           & 84.02$\pm$0.35                     & 58.92$\pm$0.03                     & 55.13$\pm$0.06                     & 51.68$\pm$0.31                     \\

\multicolumn{5}{l}{}                                                                                                                                  \\
\multicolumn{5}{l}{\textbf{CIFAR10 ($\epsilon=12/255$)}}                                                                                               \\ \hline
\multicolumn{1}{l|}{AT}               & 82.46$\pm$0.17                     & 47.73$\pm$0.08                     & 29.97$\pm$0.04                     & 27.98$\pm$0.15                     \\ 

\multicolumn{1}{r|}{(ES)}           & 78.29$\pm$0.05                     & 49.75$\pm$0.41                     & 39.77$\pm$0.35                     & 35.82$\pm$0.17                     \\ 

\multicolumn{1}{l|}{TRADES}            & 81.06$\pm$0.59                     & 50.59$\pm$0.19                     & 33.86$\pm$1.44                     & 25.21$\pm$1.92                     \\ 
            
\multicolumn{1}{r|}{(ES)}           & 78.17$\pm$2.02                     & 48.00$\pm$2.03                     & 37.27$\pm$0.65                     & 30.35$\pm$3.30                     \\ 

\multicolumn{1}{l|}{MART}               & 79.11$\pm$0.23                     & 51.23$\pm$0.16                     & 37.86$\pm$0.85                     & 31.74$\pm$0.56                     \\

\multicolumn{1}{r|}{(ES)}           & 72.04$\pm$0.32                     & 48.45$\pm$1.03                     & 41.82$\pm$0.31                     & 34.09$\pm$0.87                     \\

\multicolumn{1}{l|}{BAT}               & 81.53$\pm$0.38                     & 48.64$\pm$0.23                     & 33.42$\pm$0.62                     & 30.73$\pm$0.42                     \\ 

\multicolumn{1}{r|}{(ES)}           & 80.65$\pm$0.04                     & 49.47$\pm$0.27                     & 41.08$\pm$0.40                     & 36.03$\pm$0.21                     \\ 

\multicolumn{5}{l}{}                                                                                                                                  \\
\multicolumn{5}{l}{\textbf{CIFAR10 ($\epsilon=16/255$)}}                                                                                               \\ \hline
\multicolumn{1}{l|}{AT}               & 79.10$\pm$0.35                     & 41.31$\pm$0.71                     & 19.05$\pm$0.03                     & 16.15$\pm$0.11                     \\ 

\multicolumn{1}{r|}{(ES)}           & 72.43$\pm$0.01                     & 42.48$\pm$0.18                     & 28.78$\pm$0.16                     & 24.23$\pm$0.53                     \\ 

\multicolumn{1}{l|}{TRADES}            & 76.69$\pm$2.60                     & 40.29$\pm$1.94                     & 23.83$\pm$6.12                     & 18.68$\pm$4.65                     \\ 

\multicolumn{1}{r|}{(ES)}           & 75.01$\pm$0.46                     & 41.30$\pm$0.28                     & 28.03$\pm$0.06                     & 21.86$\pm$0.18                     \\ 

\multicolumn{1}{l|}{MART}               & 76.04$\pm$0.00                     & 45.14$\pm$0.18                     & 26.53$\pm$0.31                     & 19.62$\pm$0.52                     \\

\multicolumn{1}{r|}{(ES)}           & 68.09$\pm$0.45                     & 41.83$\pm$0.13                     & 32.61$\pm$0.40                     & 23.36$\pm$0.19                     \\ 

\multicolumn{1}{l|}{BAT}               & 78.59$\pm$0.43                     & 42.12$\pm$0.16                     & 23.61$\pm$0.43                     & 19.95$\pm$0.48                     \\ 

\multicolumn{1}{r|}{(ES)}           & 76.06$\pm$0.01                     & 41.03$\pm$0.43                     & 30.91$\pm$0.35                     & 25.06$\pm$0.34                     \\

\end{tabular}
\end{table}

For more reliable verification, we also perform an evaluation with step-wise learning rate decay. The learning rate is divided by 10 at epochs 40, 60, and 80. Furthermore, considering the recent work that uncovered the overfitting phenomenon in adversarial training \cite{rice2020overfitting}, we select the best checkpoint by PGD$^{10}$ accuracy on the first batch of the test set. We summarize the performance of the final model and the best checkpoint model. We denote the best checkpoint model with early stopping as ES. As in \Tabref{table:whitebox_cifar10}, almost all methods show improved performance against AutoAttack. TRADES with $\epsilon=8/255$ is the only case which shows accuracy drop against AutoAttack. We presume that this is caused by using PGD accuracy to early stopping not AutoAttack.

TRADES shows better performance than AT without early stopping. This result is consistent with the results of the recent work \cite{rice2020overfitting}. MART achieves the best robustness against PGD$^{50}$. However, against AutoAttack, MART shows a large decrease in robustness. Here, we note that MART seems to be overfitted to PGD. For $\epsilon=8/255$, MART shows 53.2\% accuracy against PGD with the best checkpoint. Nevertheless, when we consider untargeted APGD$_\text{DLR}$ and targeted APGD$_\text{DLR}$ \cite{croce2020reliable}, the robustness decreases to 49.6\% and 47.9\%, respectively. This tendency becomes progressively worse as $\epsilon$ increases. The proposed model shows the best robust accuracy against AutoAttack as shown in \Tabref{table:whitebox_cifar10}.
Especially, for $\epsilon=12/255$ and $16/255$, the proposed method achieves the highest accuracy not only on the robustness but also on the standard accuracy.

\paragraph{Different network architecture.}

\begin{table}
\caption{Robustness accuracy (\%) on CIFAR10 with ResNet-18.}
\label{table:diff_arc}
\centering
\begin{tabular}{lcccg}
\multicolumn{1}{l|}{Method} & \multicolumn{1}{c}{Clean}                
& \multicolumn{1}{c}{FGSM}                    
& \multicolumn{1}{c}{PGD$^{50}$}                    
& \multicolumn{1}{g}{AutoAttack}                      \\ \hline

\multicolumn{5}{l}{}                                                                                                                                                                              \\
\multicolumn{5}{l}{\textbf{CIFAR10 ($\epsilon=8/255$)}}                                                                                                                                          \\ \hline
\multicolumn{1}{l|}{AT}    & 82.36                                       & \textbf{57.48}                           & 51.87                  & 48.06                     \\
\multicolumn{1}{l|}{TRADES} & 80.91                                       & 55.45                                     & 51.94                  & 47.88                     \\
\multicolumn{1}{l|}{BAT}    & \textbf{83.56}                              & 56.27                                       & \textbf{52.20}          & \textbf{48.44}                    \\
\multicolumn{5}{l}{}                                                                                                                                                                              \\
\multicolumn{5}{l}{\textbf{CIFAR10 ($\epsilon=16/255$)}}                                                                                                                                          \\ \hline
\multicolumn{1}{l|}{AT}    & 69.75                                       & \textbf{40.17}                              & 29.47                   & 23.02                    \\
\multicolumn{1}{l|}{TRADES} & 68.26                                       & 35.23                                      & 26.13                  & 18.01                     \\
\multicolumn{1}{l|}{BAT}    & \textbf{77.02}                              & 36.64                                       & 27.22         & 21.78           \\
\multicolumn{1}{l|}{BAT ($\beta=10)$}    & 73.17                              & 37.73                                       & \textbf{29.98}         & \textbf{23.65}           
\end{tabular}
\end{table}

\Tabref{table:diff_arc} shows the results with different network architecture, ResNet-18 \cite{he2016deep}. We use the same setting in \Secref{sec:Experiments} for other settings. Similar to the results with WRN-28-10, the proposed method shows the best performance on clean examples and AutoAttack adversarial examples. For $\epsilon=16/255$, the proposed method with $\beta=5$ (BAT) achieves the highest standard accuracy with 77.02\% which is overwhelmingly higher than other methods. However, it shows insufficient robustness against AutoAttack. Thus, we also test the proposed method  with $\beta=10$ (BAT ($\beta=10$)), and it achieves the best robustness among the all methods. In addition, except the proposed method with $\beta=5$, the proposed method with $\beta=10$ also achieves the best standard accuracy.

\subsection{Experiment on CIFAR100}
For CIFAR100, we use ResNet-18 \cite{he2016deep}. We train the network for 100 epochs with SGD with an initial learning rate of 0.1, momentum of 0.9, and weight decay of $5\times10^{-4}$. We use $\epsilon=8/255$ and PGD$^{10}$ to generate adversarial examples in the training session with a step-size of $2/255$. Similar to CIFAR10, we use the best checkpoint. We use step-wise learning rate decay divided by 10 at epochs 100 and 150.
Horizontal flip and cropping are used for data augmentation. 
The robustness regularization hyper-parameter is set to $\beta=5$ for TRADES, MART. For the proposed method, we choose the best $\beta$ over $\{1,2,3,5,10\}$.

\begin{table}
\caption{Robustness accuracy (\%) on CIFAR100.}
\label{table:whitebox_cifar100}
\centering
\begin{tabular}{lcccg}
\multicolumn{1}{l|}{Method} & \multicolumn{1}{c}{Clean}                
& \multicolumn{1}{c}{FGSM}                    
& \multicolumn{1}{c}{PGD$^{50}$}                    
& \multicolumn{1}{g}{AutoAttack}                      \\ \hline

\multicolumn{5}{l}{}                                                                                                                                                                             \\ \hline
\multicolumn{1}{l|}{AT}    & \textbf{55.92}                                       & \textbf{30.13}                              & 26.51                  & 23.70                     \\
\multicolumn{1}{l|}{TRADES} & 54.56                                       & 29.82                                      & 26.48                  & 23.28                     \\
\multicolumn{1}{l|}{BAT}    & 53.67                              & 29.96                                       & \textbf{27.76}          & \textbf{23.98}                    \\
\end{tabular}
\end{table}

As shown in \Tabref{table:whitebox_cifar100}, the proposed method achieves highest robustness against PGD$^{50}$ and AutoAttack. However, in contrast to the result on CIFAR10, the proposed method could not achieve a better standard accuracy. We expect that it is difficult to satisfy Assumption \ref{asu:gamma} for CIFAR100, because it has a higher output dimension. This may be a possible reason for the result with the proposed method.

\end{document}